%% file: paper.tex
 \documentclass[pmlr,twocolumn,10pt]{jmlr} % W&CP article

\usepackage{booktabs}
\usepackage{siunitx}
\usepackage{rotating}

\usepackage[switch]{lineno}

\theorembodyfont{\upshape}
\theoremheaderfont{\scshape}
\theorempostheader{:}
\theoremsep{\newline}

\jmlrvolume{297}
\jmlryear{2025}
\jmlrworkshop{Machine Learning for Health (ML4H) 2025} % W&CP title

 \title[\textsc{bebms}]{Bayesian Event-Based Model for Disease Subtype and Stage Inference}

\author{%
\Name{Hongtao Hao}\Email{hhao9@wisc.edu}\\
\addr University of Wisconsin–Madison, USA
\AND
\Name{Joseph L. Austerweil }\Email{joseph.austerweil@gmail.com}\\
\addr \addr Chiba Institute of Technology, Japan \& University of Wisconsin-Madison, USA
\AND
\Name{for the Alzheimer’s Disease Neuroimaging Initiative\textsuperscript{*}}
}

\begin{document}

\maketitle

\begingroup
\renewcommand\thefootnote{*}
\footnotetext{
Data used in preparation of this article were obtained from the Alzheimer’s Disease Neuroimaging Initiative (ADNI) database (\url{adni.loni.usc.edu}). As such, the investigators within the ADNI contributed to the design and implementation of ADNI and/or provided data but did not participate in analysis or writing of this report. A complete listing of ADNI
investigators can be found in Appendix \ref{sec:adni}.}
\endgroup

\begin{abstract}
Chronic diseases often progress differently across patients. Rather than randomly varying, there are typically a small number of subtypes for how a disease progresses across patients. To capture this structured heterogeneity, the Subtype and Stage Inference Event-Based Model (SuStaIn) estimates the number of subtypes, the order of disease progression for each subtype, and assigns each patient to a subtype from primarily cross-sectional data. It has been widely applied to uncover the subtypes of many diseases and inform our understanding of them. But how robust is its performance? In this paper, we develop a principled Bayesian subtype variant of the event-based model (\textsc{bebms}) and compare its performance to SuStaIn in a variety of synthetic data experiments with varied levels of model misspecification. BebmS substantially outperforms SuStaIn across ordering, staging, and subtype assignment tasks. Further, we apply \textsc{bebms} and SuStaIn to a real-world Alzheimer's data set. We find \textsc{BebmS} has results that are more consistent with the scientific consensus of Alzheimer's disease progression than SuStaIn.    
\end{abstract}
\begin{keywords}
Event-based model, Disease progression, Bayesian methods, Subtypes, Alzheimer's disease
\end{keywords}

\paragraph*{Data and Code Availability}
\textsc{bebms} can be installed by \texttt{pip install bebms} (\url{https://github.com/jpcca/bebms_pkg}). After installing the \textsc{bebms} package, the codes necessary for replicating the experiments using High-Throughput Computing (HTC) on top of a cluster environment are available at \url{https://github.com/hongtaoh/bebms}. ADNI data can be requested at \url{https://adni.loni.usc.edu}.

\paragraph*{Institutional Review Board (IRB)} 
The IRB at University of Wisconsin-Madison has reviewed and approved the research (\#2025-1254).

\section{Introduction}
\label{sec:intro}
\input{1_introduction}

\section{Past Work}
\input{2_literature}

\section{Method}
\input{3_methods}

\section{Model Evaluation}
\label{main:model_evaluation}
\input{4_experiments}

\section{Results}
\input{5_results}

\section{Discussion}
\input{6-discussion}

% ACKNOWLEDGEMENTS ONLY GO IN THE CAMERA-READY, NOT THE SUBMISSION
\acks{

We thank the CHTC at the University of Wisconsin-Madison for computing support. JLA was funded by the Japan Probabilistic Computing Consortium Association (JPCCA).

Data collection and sharing for this project was funded by the Alzheimer's Disease
Neuroimaging Initiative (ADNI) (National Institutes of Health Grant U01 AG024904) and
DOD ADNI (Department of Defense award number W81XWH-12-2-0012). ADNI is funded
by the National Institute on Aging, the National Institute of Biomedical Imaging and
Bioengineering, and through generous contributions from the following: AbbVie, Alzheimer’s
Association; Alzheimer’s Drug Discovery Foundation; Araclon Biotech; BioClinica, Inc.;
Biogen; Bristol-Myers Squibb Company; CereSpir, Inc.; Cogstate; Eisai Inc.; Elan
Pharmaceuticals, Inc.; Eli Lilly and Company; EuroImmun; F. Hoffmann-La Roche Ltd and
its affiliated company Genentech, Inc.; Fujirebio; GE Healthcare; IXICO Ltd.; Janssen
Alzheimer Immunotherapy Research \& Development, LLC.; Johnson \& Johnson
Pharmaceutical Research \& Development LLC.; Lumosity; Lundbeck; Merck \& Co., Inc.;
Meso Scale Diagnostics, LLC.; NeuroRx Research; Neurotrack Technologies; Novartis
Pharmaceuticals Corporation; Pfizer Inc.; Piramal Imaging; Servier; Takeda Pharmaceutical
Company; and Transition Therapeutics. The Canadian Institutes of Health Research is
providing funds to support ADNI clinical sites in Canada. Private sector contributions are
facilitated by the Foundation for the National Institutes of Health (www.fnih.org). The grantee
organization is the Northern California Institute for Research and Education, and the study is
coordinated by the Alzheimer’s Therapeutic Research Institute at the University of Southern
California. ADNI data are disseminated by the Laboratory for Neuro Imaging at the
University of Southern California.

% Acknowledgments to ADNI can be found in Appendix \ref{sec:adni}.
}

\bibliography{paper}

% Switch to one column before appendix
\onecolumn
\appendix
\input{appendix}

\end{document}

%% file: 1_introduction.tex
Understanding how chronic diseases progress is crucial for early diagnosis, prognosis, and therapeutic development~\citep{jack2010hypothetical}. These diseases rarely unfold along a single pathway. In the case of Alzheimer's disease, patients often follow distinct trajectories characterized by different progression sequences~\citep{estarellas2024multimodal, vogel2021four, ten2018atrophy, poulakis2022multi, jellinger2021pathobiological}. Identifying subtypes is critical for uncovering disease mechanisms, and personalizing medicine via improved diagnoses and tailored interventions.

Progression modeling frameworks, such as the Event-Based Model \citep[EBM;][]{fonteijn2012event}, are powerful tools for reconstructing disease progression from cross-sectional data. However, most EBM variants assume a canonical trajectory, limiting their ability to capture heterogeneity. One exception is the Subtype and Stage Inference Event-Based Model \citep[SuStaIn;][]{young2018uncovering, aksman2021pysustain}. It addressed this limitation by extending EBM to multiple subtypes, and it has since become the de facto standard, applied to a wide range of neurodegenerative diseases in high-profile publications~\citep{vogel2021four, salvado2024disease, estarellas2024multimodal, eshaghi2021identifying, mastenbroek2024disease}. Despite this impact, SuStaIn has not been rigorously evaluated for robustness, particularly under model misspecification (e.g., non-Gaussian biomarker distributions, continuous disease stages, and uneven subtype and stage distributions). We investigated its performance on realistic synthetic datasets with known ground truth and its performance was brittle.

In this paper, we introduce the Bayesian Event-Based Model for Subtyping (\textsc{bebms}). Our approach retains the interpretability of EBMs while embedding them in a Bayesian framework, enabling more accurate inference of biomarker orderings, disease staging, and subtype assignment. Across diverse realistic synthetic datasets, we show that \textsc{bebms} improves biomarker ordering (27\%), disease staging (89\%), and subtype assignment (56\%) over SuStaIn, while reducing runtime.

%% file: 2_literature.tex
The EBM~\citep{fonteijn2012event} formulates disease progression as a sequence of biomarker events, where each biomarker switches from a healthy to a pathological distribution at an unknown position in the sequence. Once the latent disease stage exceeds this position, the biomarker is considered affected. This framework enabled estimation of progression patterns in several neurodegenerative diseases from primarily cross-sectional data~\citep{young2018uncovering, fonteijn2012event, oxtoby2021sequence, wijeratne2023temporal, firth2020sequences}. 

Over time, a number of extensions to EBM have been proposed. The Discriminative EBM~\citep[DEBM;][]{venkatraghavan2019disease} relaxed the assumption of one ordering by allowing subject-specific variability as random Mallows-distributed noise around the canonical ordering. The Temporal EBM \citep[TEBM;][]{wijeratne2023temporal} reformulated progression in continuous rather than discrete time. The Parsimonious EBM \citep[P-EBM;][]{csparsimonious} captures cases where multiple biomarkers become pathological simultaneously. The KDE EBM \citep{firth2020sequences} introduced nonparametric likelihoods, enabling the model to estimate data likelihood under non-Gaussian biomarker distributions. The Stage Aware EBM \citep[SA-EBM;][]{hao2025stage} introduced stage distributions and improved inference, resulting in substantial performance improvements over the original EBM, KDE EBM, and DEBM. Despite these advances, all of these models assume a single central ordering, and therefore cannot capture disease heterogeneity across patients.

SuStaIn~\citep{young2018uncovering} addressed this limitation by extending EBM to incorporate multiple progression patterns, enabling subtype inference. SuStaIn has been widely adopted across Alzheimer's disease~\citep{vogel2021four, salvado2024disease, estarellas2024multimodal}, multiple sclerosis~\citep{eshaghi2021identifying}, and Lewy body disease~\citep{ mastenbroek2024disease}, and is supported by an open-source implementation~\citep{aksman2021pysustain}. As SuStaIn is computationally demanding, scalable variants for high-dimensional data~\citep[s-SuStaIn;][]{tandon2024s} and incomplete data~\citep{estarellas2024multimodal} have been developed. 

While SuStaIn and its extensions represent a major step forward in capturing disease heterogeneity, they also have important limitations. First, they typically assume a uniform prior over disease stages, even though later stages are underrepresented in real-world cohorts such as ADNI~\citep{donohue2014estimating}. Second, they rely on static estimates of biomarker distribution parameters, which remain fixed during inference of subtype orderings and may bias results when the true ordering and stage distribution are unknown. Despite its widespread adoption, the robustness of SuStaIn has rarely been systematically evaluated on synthetic datasets with known ground truth. As a result, the model's robustness under misspecification remains unclear. 

Building on prior work, we introduce the Bayesian Event-Based Model for Subtyping (\textsc{bebms}). By embedding EBMs in a Bayesian framework, \textsc{bebms} provides more accurate estimation of disease subtypes, biomarker orderings, and patient stages, while also reducing runtime, as demonstrated in both synthetic and ADNI experiments.

%% file: 3_methods.tex
\subsection{Model Specification}
\begin{figure}[htbp]
\floatconts
  {fig:bebmsgraphmod}
  {\vspace{-0.5cm}\caption{\textsc{bebms} as a graphical model.
  }}
  {\includegraphics[width=0.975\linewidth]{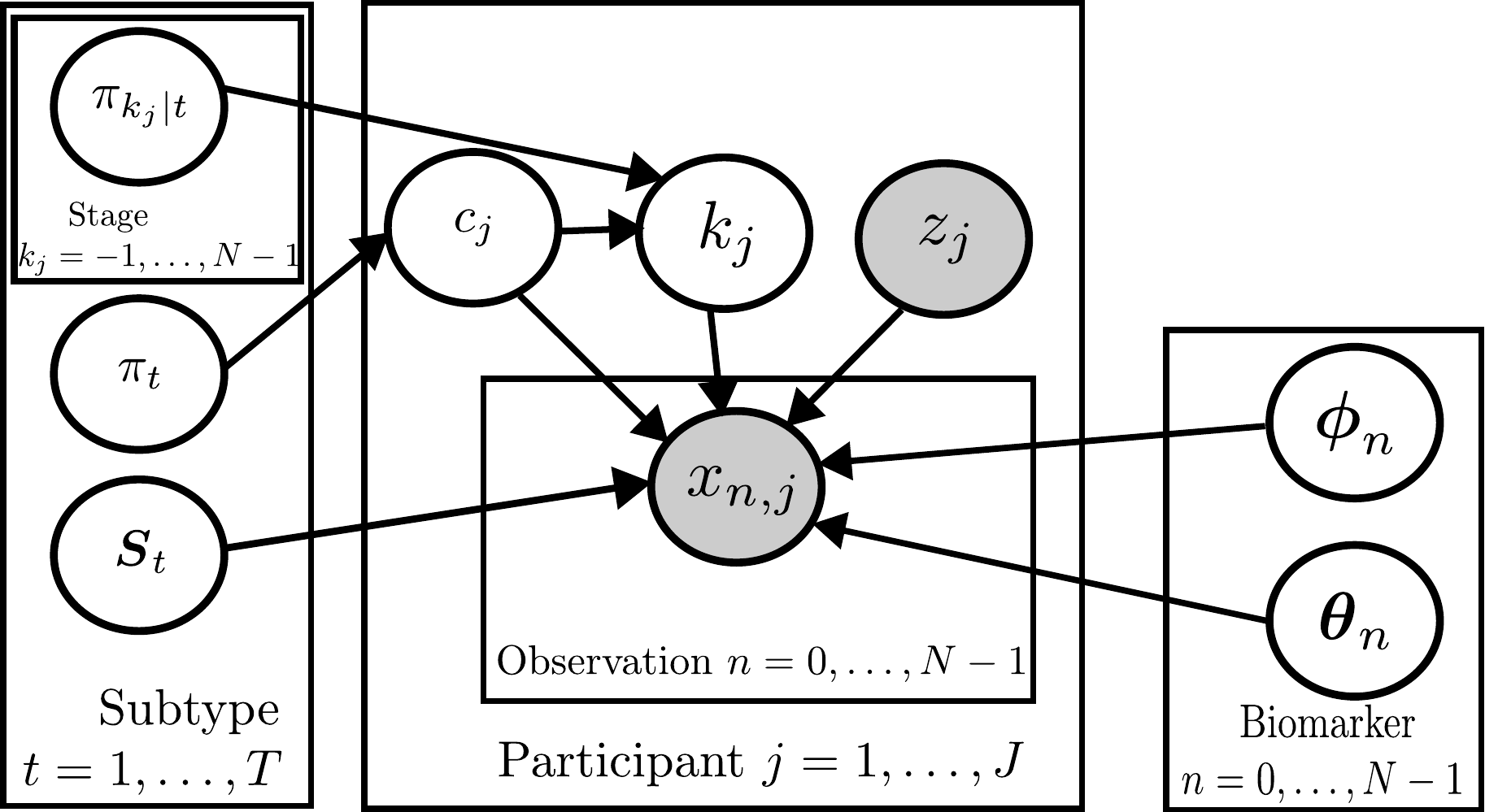}}
\end{figure}

\textsc{bebms} models disease progression as a sequence of biomarker events. Each of the $N$ biomarkers can exist in either a ``pre-event'' (healthy) or ``post-event'' (pathological) state. A participant $j$'s disease state is defined by their stage $k_j$. We define $k_j = -1$ for a healthy participant (no events), and $k_j \in [0, 1, ..., N-1]$ for a progressing participant, where $k_j$ corresponds to the highest rank of the event that has occurred. 

We model $T$ disease subtypes, each defined by a unique sequence of biomarker events. Let $\boldsymbol{S}$ be a $T \times N$ matrix where $S_{t,n}$ is the 0-based \textbf{rank} of biomarker $n$ in the progression sequence for subtype $t$. We also write $S(t,i)$ for the \textbf{biomarker} at position $i$ in subtype $t$, so $S(t,i) = n$ if and only if $S_{t,n}=i$. A biomarker $n$ is considered post-event for a participant in subtype $t$ and stage $k_j$ if $k_j \ge S_{t,n}$. Let $c_j$ denote participant $j$'s subtype. $\pi_{t}$ and $\pi_{k_j | t}$ reflect the probabilities of subtype $t$ ($c_j \sim \pi_{t}$) and of a participant being in disease stage $k_j$ given they belong to subtype $t$ ($k_j |c_j \sim \pi_{t|c_j}$, respectively. Figure ~\ref{fig:bebmsgraphmod} presents the approach as a graphical model.

Biomarker measurements in both states are modeled with Gaussian distributions. \textbf{We assume a shared parameterization for biomarkers across subtypes}:

\begin{equation}
x_{j,n} \sim 
\begin{cases}
\mathcal{N}(\phi_{n,\mu}, \phi_{n,\sigma}^2), & \text{if pre-event}, \\
\mathcal{N}(\theta_{n,\mu}, \theta_{n,\sigma}^2), & \text{if post-event}.
\end{cases}
\end{equation}

\noindent with $\phi$ and $\theta$ denoting healthy and afflicted distribution parameters, respectively. $\boldsymbol{X}$ is the whole dataset, and $\boldsymbol{X}_j$ is the biomarker measurements of a specific participant. We use $x_{j,n}$ to denote the value of biomarker $n$ in participant $j$.

The model is a mixture over subtypes and stages. For a healthy participant  ($z_j = 0$, i.e., $k_j = -1$), the data likelihood is 

\begin{equation}
p(\boldsymbol{X}_j \mid \boldsymbol{S}, z_j = 0) = \prod_{n=0}^{N-1} p\!\left(x_{j,n} \mid \phi_{n}\right)
\end{equation}

For a diseased participant ($z_j = 1$), the likelihood is marginalized over all subtypes $t$ and all possible disease stages $k_j$ for participant $j$, weighted by their respective priors $\pi_t$ and $\pi_{k_j\mid t}$:

\begin{equation}
p(\boldsymbol{X}_j \mid \boldsymbol{S}, z_j = 1) =\sum_{t=1}^T \pi_t \sum_{k_j=0}^{N-1} \pi_{k_j \mid t} \ p(\boldsymbol{X}_j\mid \boldsymbol{S}_t, z_j, k_j)
\end{equation}

\noindent where $p(\boldsymbol{X}_j \mid \boldsymbol{S}_t, z_j = 1, k_j)$ denotes the likelihood of participant $j$ given they are in subtype $t$ at stage $k_j$:

\begin{align}
p(\boldsymbol{X}_j \mid \boldsymbol{S}_t, z_j{=}1, k_j)
&= 
\underbrace{\prod_{i=0}^{k_j}
  p(x_{j,S(t,i)} \mid \theta_{S(t,i)})}_{\text{post-event}}
  \notag\\[-3pt]
&\quad\times
\underbrace{\prod_{i=k_j+1}^{N-1}
  p(x_{j,S(t,i)} \mid \phi_{S(t,i)})}_{\text{pre-event}}.
\end{align}

The total data likelihood across all $J$ participants is the product of their individual likelihoods:

\begin{equation}
    P(\boldsymbol{X}\mid \boldsymbol{S}) = \prod_{j=1}^J  P(\boldsymbol{X}_j \mid \boldsymbol{S}, z_j)
\end{equation}

We place Dirichlet priors on the subtype prior $\boldsymbol{\pi} \in \mathbb{R}^T \sim \text{Dir}(\boldsymbol{\alpha})$ and the stage prior $\boldsymbol{\pi}_{\cdot \mid t} \in \mathbb{R}^{T \times N} \sim \text{Dir}(\boldsymbol{\alpha}_{\cdot \mid t})$, with weakly informative priors~\citep{gelman2017prior} of $1.0$ for $\boldsymbol{\alpha}$ and $\boldsymbol{\alpha}_{\cdot \mid t}$.

\subsection{Inference Procedure}

The biomarker parameters ($\boldsymbol{\theta}, \boldsymbol{\phi}$) are initialized by K-Means using controls for pre-event and progressing participants for post-event. After clustering, the cluster containing the majority of controls is labeled pre-event ($\boldsymbol{\phi}$) and the other post-event ($\boldsymbol{\theta}$). K-Means estimates are immediately refined with weighted conjugate updates. We assume a Normal-Inverse Gamma (NIG) conjugate prior on the unknown mean and variance, $(\mu, \sigma^2)\sim \text{NIG}(m_0, n_0, s^2_0, \nu_0)$. The posterior is also a NIG distribution with updated parameters. 

We set $n_0 = 1, \nu_0=1$, in the spirit of weakly informative priors~\citep{gelman2017prior}. $m_0$ and $s_0^2$ are the raw mean and variance of each cluster, respectively. The updates use a soft assignment of each measurement to the pre- and post-event clusters. Each cluster has a weight vector ($\boldsymbol{w} \in \mathbb{R}^J$), indicating the posterior probability of each subject belonging to the cluster. For initialization, we assign 1 to all entries of $\boldsymbol{w}$. See Appendix~\ref{apd:cp} for how we update distribution parameters using conjugate priors.

A key innovation of \textsc{bebms} is the iterative estimation of distribution parameters ($\boldsymbol{\theta}, \boldsymbol{\phi}$), and stage ($\boldsymbol{\pi}_{\cdot \mid t}$) and subtype ($\boldsymbol{\pi}$) priors through a Metropolis-Hastings Markov chain Monte Carlo (MCMC) sampler (See Algorithm~\ref{algo:mcmc} in Appendix~\ref{apd:inference_algo}).

Specifically, we compute the intermediate stage ($\tilde{P}_{\text{stage}} \in \mathbb{R}^{J \times T \times N}$) and subtype ($\tilde{P}_{\text{subtype}} \in \mathbb{R}^{J \times T}$) posteriors as follows:

$$P_\text{stage}(k\mid j, t) = \frac{\pi_{k\mid t} \cdot p(\boldsymbol{X}_j\mid \boldsymbol{S}_t, k, \boldsymbol{\theta}, \boldsymbol{\phi})}{\sum_{k^\prime=0}^{N-1}\pi_{k^\prime\mid t} \cdot p(\boldsymbol{X}_j\mid \boldsymbol{S}_t, k^\prime, \boldsymbol{\theta}, \boldsymbol{\phi})}$$

$$P_{\text{subtype}}(t\mid j) = \frac{\pi_t \sum_{k=0}^{N-1}P_{\text{stage}}(k \mid j, t)}{\sum_{t^\prime = 1}^T\pi_{t^\prime} \sum_{k=0}^{N-1}\pi_{k \mid t^\prime} P_{\text{stage}}(k \mid j, t^\prime)}$$

In updating distribution parameters, the weights ($\boldsymbol{w}$) are the marginalized probabilities of being in pre/post-event states across all subtypes and stages:

$$w_{j, n, \theta} = \sum_{t=1}^T \sum_{k=0}^{N-1} \mathbf{1}_{\{k \ge S_{t,n}\}} P_{\text{stage}}(k\mid j, t) P_{\text{subtype}}(t \mid j),$$

$$w_{j, n, \phi} = 1 - w_{j, n, \theta}$$

\noindent For healthy participants, assignments are fixed: $w_{j, n, \phi} = 1, w_{j, n, \theta}=0$ for all biomarkers. If a proposal is accepted, we update subtype and stage priors with posterior counts:

$$\boldsymbol{\pi} \sim \text{Dir}\left( \boldsymbol{\alpha} + \sum_{j=1}^J P_{\text{subtype}}(t \mid j) \right)$$

$$\boldsymbol{\pi}_{\cdot \mid t} \sim \text{Dir}\left(\boldsymbol{\alpha}_{\cdot \mid t} + \sum_{j=1}^J P_{\text{stage}}(k \mid j, t) P_{\text{subtype}}(t \mid j)\right)$$

\subsection{Model Selection}

The above inference procedure assumes we know the number of subtypes $T$. In real-world scenarios, however, this information is missing. Following SuStaIn~\citep{aksman2021pysustain,young2018uncovering}, we applied $K$-fold cross-validation to find the optimal $T$. For each candidate $T$, we trained the model on the training folds and obtained the out-of-sample log-likelihood on the held-out fold. We then aggregated log-likelihoods across folds to compute a cross-validation information criterion (CVIC), and selected the $T$ with the lowest CVIC score. When multiple $T$ have similar CVIC scores (difference less than 6), we chose the smallest $T$ within the group. Specifically, we performed stratified $K$-fold cross-validation to maintain the proportion of progressing/healthy in each fold. CVIC score is computed as:

\begin{equation}
    \text{CVIC} = 2 \cdot \sum_{i=1}^T i\text{th fold log likelihood}
\end{equation}

\noindent We chose the threshold of 6 for consistency with SuStaIn's procedure~\citep{aksman2021pysustain}.

%% file: 4_experiments.tex
We use SuStaIn as the baseline for evaluation because it is the only publicly available implementation to reconstruct subtypes of disease progression using cross-sectional data. Missing data SuStaIn~\citep{estarellas2024multimodal} referred to SuStaIn as the code source and s-SuStaIn~\citep{tandon2024s} does not have publicly available code. For the single-subtype case, \textsc{bebms} is essentially the same as the Conjugate Prior variant of SA-EBM~\citep{hao2025stage}, which has improved performance compared to DEBM \citep{venkatraghavan2019disease}, KDE-EBM \citep{firth2020sequences} and UCL GMM~\citep{firth2020sequences}. Thus, SuStaIn is the only benchmark algorithm in this study. 

Our evaluation relies on both synthetic and real-world datasets. The real-world data comes from the Alzheimer's Disease Neuroimaging Initiative \citep[ADNI,][]{mueller2005alzheimer}. The ADNI was launched in 2003 as a public-private partnership, led by Principal Investigator Michael W. Weiner, MD. The primary goal of ADNI has been to test whether serial magnetic resonance imaging (MRI), positron emission tomography (PET), other biological markers, and clinical and neuropsychological assessment can be combined to measure the progression of mild cognitive impairment (MCI) and early Alzheimer’s disease (AD). 

Our analysis was based on the \texttt{adnimerge} table (updated on September 7, 2023) from the Alzheimer's Disease Cooperative Study data system. We restrict the study cohort to baseline visits from participants with a diagnosis of CN, Early and Late MCI, or AD. The biomarker selection consists of 12 measures commonly used in the field~\citep{csparsimonious, young2014data, archetti2019multi}. These biomarkers cover a wide range of categories: cognitive assessment, Cerebrospinal fluid (CSF) markers and MRI-derived measurements of the brain regions. See Table~\ref{tab:biomarker_glossary} and~\ref{tab:biomarker_params_and_irr} (in Appendix~\ref{apd:glossary} and~\ref{apd:biomarker_params_and_irr}, respectively) for details about these biomarkers.

For the brain regions MRI measurements, following best practices, we applied intracranial volume (ICV) normalization because people's brain sizes vary. We excluded participants with missing values from any of these 12 biomarkers, and de-duplicated the final dataset. The final version of ADNI fitting these criteria had 726 participants, distributed from three ADNI protocols: ADNI (275, 37.9\%), ADNIGO (76, 10.5\%) and ADNI2 (375, 51.7\%). There were 413 (56.9\%) men and 313 (43.1\%) women, diagnosed as AD (153, 21.1\%), late MCI (236, 32.5\%), Control (155, 21.3\%) and early MCI (182, 25.1\%). 

We obtained the biomarker distribution parameters for synthetic datasets based on ADNI. To provide SuStaIn with the best opportunity to perform well, we applied their method (Gaussian Mixture Model) on the processed ADNI dataset to obtain $\boldsymbol{\theta}$ and $\boldsymbol{\phi}$.

We employed two models to generate synthetic datasets: the EBM and the Sigmoid model (defined later). First, we uniformly pick a number from 1 to 5 (inclusive) for the number of subtypes. We chose [1,5] because [2,5] is the most likely range for AD according to the literature~\citep{young2018uncovering, estarellas2024multimodal, ten2018atrophy, jellinger2021pathobiological}, and $1$ to account for no subtypes. We then randomly pick a dispersion parameter between 0.01 and 0.5, and use the Top-K Mallow's Model implemented by~\citet{topkMallows} to get the event sequence for each subtype based on a random permutation of all the 12 biomarkers. We chose [0.01, 0.5] because it allows the generated subtypes' sequences to have varied agreement covering the whole range of [0,1] as measured by Kendall's $W$ (See Appendix~\ref{apd:kendalls_w}). We made sure no two subtypes share the same event sequence, and there were at least ten progressing participants in each subtype. 

Healthy participants do not belong to any subtype. We used a Dirichlet-Multinomial (DM) distribution to assign subtypes to progressing participants. The prior for the DM distribution is selected uniformly at random from 0.1, 2, 5, and 20. This allows both sparse and uniform distributions for participants' subtype assignments. DM distribution is also employed to generate disease stages for progressing participants, but the prior setting depends on experimental configurations (See below and Appendix~\ref{apd:experiment_specifications}).

Given the event sequence ($\boldsymbol{S}_t$) of subtype $t$, a biomarker $n$, and a participant $j$ with diagnosis $z_j \in \{0,1 \}$ and disease stage $k_j$, the generative model of EBM defines the biomarker measurement of $x_{j,n} \mid \boldsymbol{S}_t, k_j, \boldsymbol{\theta}_n, \boldsymbol{\phi}_n, z_j$ as:

\begin{equation} \label{eq:gen}
x_{j,n} \sim 
\begin{cases}
p(x_{n,j} \mid \boldsymbol{\theta}_n), & z_j = 1, \; S_{t,n} \leq k_j, \\
p(x_{n,j} \mid \boldsymbol{\phi}_n), & z_j = 1, \; S_{t,n} > k_j, \\
p(x_{n,j} \mid \boldsymbol{\phi}_n), & z_j = 0.
\end{cases}
\end{equation}

The Sigmoid model, adapted from \citet{young2015simulation} and \citet{venkatraghavan2019disease}, is motivated by the biomarker cascade hypothesis of \citet{jack2010hypothetical}, which postulates that AD biomarkers follow sigmoid-shaped trajectories: slow to change in early stages, accelerating during symptomatic onset, and plateauing later. As in EBM, measurements of healthy individuals are drawn from normal distributions. Formally:

$$x_{j,n} \sim \mathcal{N}(\mu_{n, \phi}, \sigma^2_{n, \phi})$$

The measurements of progressing participants monotonically deviate from the healthy state:

$$x_{n,j} \sim \mathcal{N}(\mu_{n, \phi}, \sigma^2_{n, \phi}) + \frac{(-1)^{I_n} R_n}{1 + e^{-\rho_n(k_j - \xi_n)}}$$

\noindent where  $I_n \sim \text{Bernoulli (0.5)}$ randomly flips the direction of the deviation, $R_n = \mu_{n, \theta} - \mu_{n, \phi}$ controls the range of the measurements, and $\rho_n = \max \left(1, \frac{|R_n|}{\sqrt{\sigma^2_{n, \theta} + \sigma^2_{n, \phi}}} \right)$ sets the slope.

SuStaIn~\citep{aksman2021pysustain} has two variants: GMM (gaussian mixture model) and KDE. GMM assumes that biomarker measurements follow Gaussian distributions whereas KDE does not. Both \textsc{bebms} and SuStaIn rely on three key assumptions: (1) biomarker measurements follow Gaussian distributions (except for SuStaIn KDE); (2) disease stages are ordinal; and (3) biomarker events occur in an ordinal sequence with approximately even spacing. To evaluate robustness, we systematically relaxed each assumption. Non-Gaussianity was tested both by generating EBM datasets with non-normal biomarker distributions (Table~\ref{tab:biomarker_params_and_irr} in Appendix~\ref{apd:biomarker_params_and_irr} has more details) and by using the Sigmoid model. Continuous disease stages were introduced to test violations of the second assumption. We examined violations of the third assumption by introducing uneven event spacing across biomarkers. Experimental results show that the last test revealed a shared limitation of the underlying modeling paradigm. See Appendix~\ref{apd:res} for more details. Details about experimental specifications are available in Appendix~\ref{apd:experiment_specifications}. In Appendix~\ref{apd:all-dist}, we have plotted the theoretical and empirical distributions of all twelve biomarkers (Fig.~\ref{fig:all-dist}).

\subsection{Experiment Setup}

For each subtype $t$, when the \textsc{bebms} infers the event sequences $\boldsymbol{S}_t$, i.e., doing the \textbf{ordering task}, it assumes knowledge of the diagnosis labels, i.e., whether healthy or progressing. In the \textbf{subtyping task} and the \textbf{staging task}, when inferring the most likely subtype and disease stage for each participant, \textsc{bebms} is blind to the diagnosis labels. It also discards the stage and subtype priors inferred by the model and relies only on the estimated biomarker distribution parameters $\boldsymbol{\theta, \phi}$.

SuStaIn's methodology differs, as it leverages diagnosis labels only to estimate the biomarker distribution parameters, excluding them from subsequent inference tasks. While this design aims for robustness against diagnostic uncertainty, it introduces a potential issue: if the labels are unreliable, their use in any estimation step risks biasing the model's core parameters. To enable a direct comparison, we developed \textsc{bebms (blind)}. This variant strictly mirrors SuStaIn's philosophy, utilizing diagnosis labels only for the initialization of biomarker distribution parameters and not in any of the three tasks mentioned above. 

We used four different total participant sizes: $J = 300, 500, 1000, 1500$ and three different healthy ratios: $R = 0.25, 0.5, 0.75$. For each $J-R$ pair, we generated 10 datasets. With eleven experiments, we have 1,320 datasets in total. \textsc{bebms} and \textsc{bebms (blind)} use 10,000 MCMC iterations with 500 burn-in (used only for visualizing uncertainties in resulting progressions, e.g., Fig.~\ref{fig:adni_ordering_bebms}) and no thinning. Per SuStaIn~\citep{aksman2021pysustain} recommendation, we applied 25 parallel start points for their E-M algorithm and 100,000 MCMC iterations. All other settings are the default of SuStaIn. We applied both the GMM and the KDE version of SuStaIn. For the estimation of $T$, in cross-validation, we tested $T=1$ to $T=5$ (inclusive). We used the maximum-likelihood ordering (the E-M solution) from SuStaIn as the estimated subtype progression pattern, and the \texttt{ml\_subtype} and \texttt{ml\_stage} outputs from \texttt{run\_sustain\_algorithm} as the subject-level subtype and stage assignments, respectively. Posterior MCMC samples (\texttt{samples\_sequence}) were used only to quantify uncertainty in the ordering results.

\subsection{Evaluation Metrics}

For the ordering task, we computed a cost matrix of normalized Kendall's $\tau$ distances and applied the Hungarian algorithm~\citep{kuhn1955hungarian, munkres1957algorithms} to optimally match estimated and true sequences, reporting the mean distance across matched pairs.  Subtype assignment accuracy was measured with the Adjusted Rand Index~\citep{hubert1985comparing} between the true and inferred subtype labels. Note that the performance on the subtyping task is only relevant when the ground truth has more than 1 subtype. As staging error is confounded with subtype accuracy, we evaluated the staging accuracy by only reporting the mean estimated stage for control participants with the ground truth of 0. We reported the Mean Absolute Error (MAE) as the accuracy for the estimation of the number of subtypes. To compare the computational efficiency, we also report the runtime for each dataset. Due to the computational cost of cross-validation, we only report the full runtime of model selection for Experiment 1.

%% file: 5_results.tex
\subsection{Synthetic Datasets}

We conducted all experiments on the CHTC cluster at the University of Wisconsin-Madison~\citep{https://doi.org/10.21231/gnt1-hw21}. 
Across the 1,320 synthetic datasets, SuStaIn failed to process 155 (155 for KDE and 1 for GMM) files due to ``IndexError'' and ``Singular matrix'', highlighting its fragility in practice. As shown in Figs.~\ref{fig:tau},~\ref{fig:subtype}, ~\ref{fig:stage} and \ref{fig:runtime}, \textsc{bebms} consistently outperforms SuStaIn across ordering, subtyping, and staging tasks, and is faster. Unless otherwise mentioned, the results refer to Exp. 1-9. 

\begin{figure*}[htbp]
\floatconts
  {fig:tau}
  {\vspace{-1cm}\caption{Normalized Kendall’s $\tau$ across all nine synthetic experiments. Each panel corresponds to an experiment; within each panel, participant sizes ($J$) are shown across columns, and within each column three healthy ratios ($R=0.25, 0.5, 0.75$) are displayed from left to right. \textsc{bebms} reduced ordering error by $27.3\%$ relative to SuStaIn, with \textsc{bebms} (Blind) performing nearly identically. SuStaIn results were consistently lower, with margins narrowing under model misspecification (Experiments 8–9). Performance was largely insensitive to participant size and healthy ratio.
  }}
  {\includegraphics[width=0.95\linewidth]{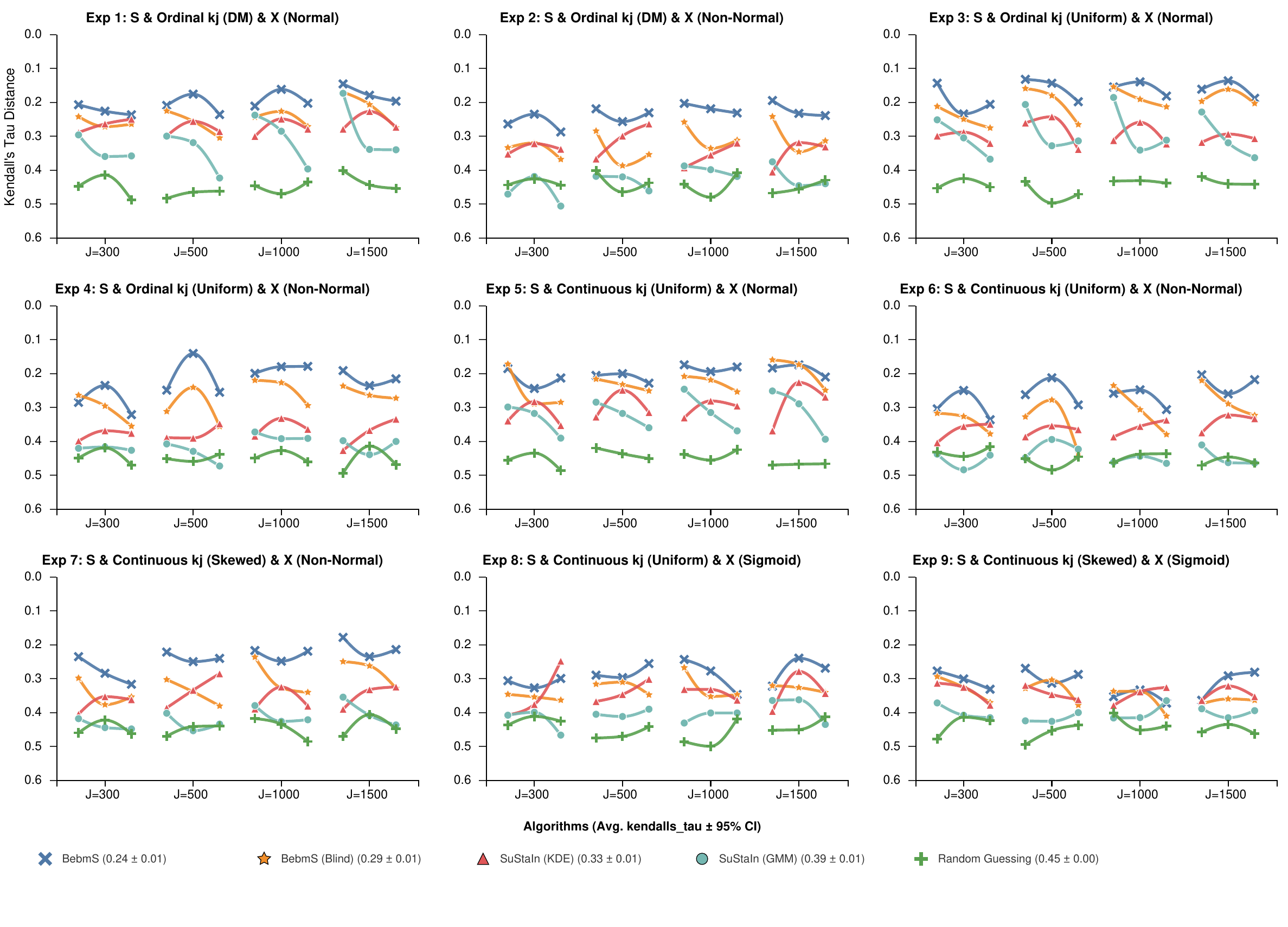}}
\end{figure*}

\begin{figure*}[htbp]
\floatconts
  {fig:adni_ordering_bebms}
  {\vspace{-0.5cm}\caption{\textsc{bebms} ADNI ordering. All subtypes begin in the entorhinal region, with Subtype 1 showing early cognitive decline, Subtype 2 early CSF changes, and Subtype 3 early neurodegeneration. 
  }}
  {\includegraphics[width=1\linewidth]{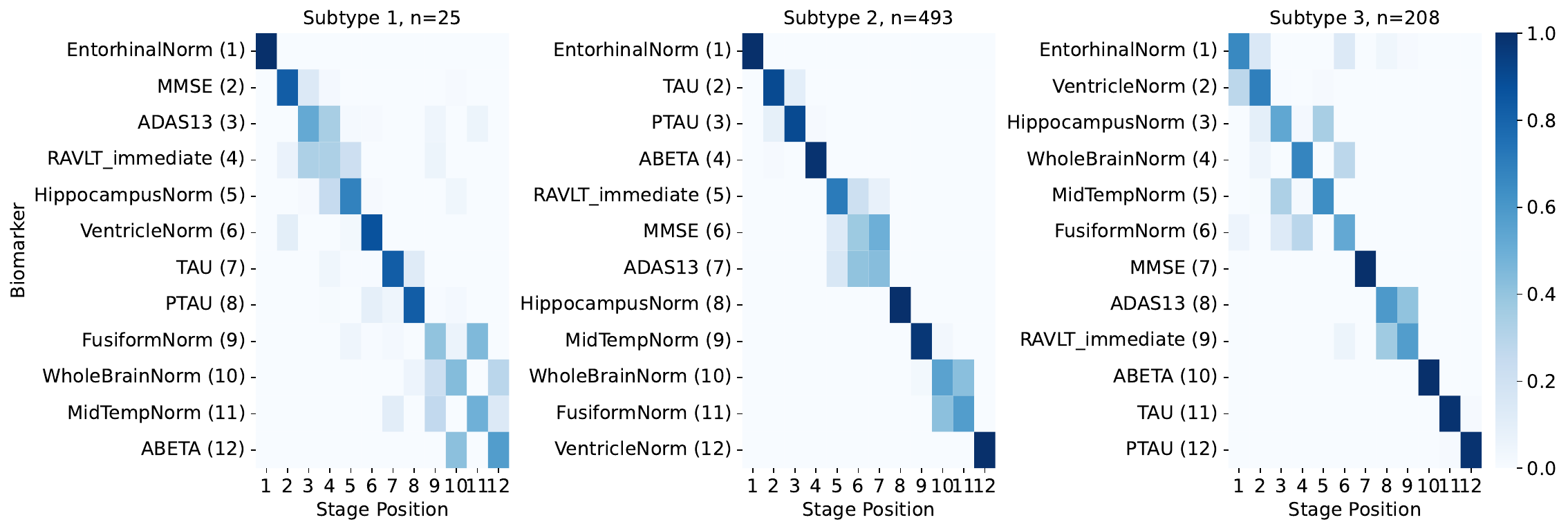}}
\end{figure*}

\textbf{Ordering:} \textsc{bebms} and its blind variant achieved Kendall’s $\tau$ distances of $0.24 \pm 0.01$ and $0.29 \pm 0.01$, respectively---substantially lower (better) than SuStaIn KDE ($0.33 \pm 0.01$) and SuStaIn GMM ($0.39 \pm 0.01$). For reference, random guessing yielded $0.45 \pm 0.00$. Variability is 95\% CI. SuStaIn GMM performed particularly poorly when biomarker measurements deviated from Gaussian assumptions and also showed a marked decline with increasing proportions of healthy participants, even under normally distributed data. By contrast, \textsc{bebms}---although also assuming Gaussian distributions---was robust to non-Gaussian misspecification, and outperformed SuStaIn KDE in most cases. Both \textsc{bebms} and SuStaIn struggled in Experiment 9, where the ordinal assumption of disease stages was violated and data were generated from the sigmoid framework. Regarding participant size, performance saturated around 300 participants. \textsc{bebms} was also robust to varying healthy ratios: holding the total number of participants fixed, increasing the proportion of controls had only a modest effect on ordering accuracy.

\textbf{Subtyping}. As shown in Fig.~\ref{fig:subtype} (Appendix~\ref{apd:res}), subtyping proved to be a challenging task overall. \textsc{bebms} achieved an Adjusted Rand Index (ARI) $0.25 \pm 0.02$, followed by \textsc{bebms (blind)} ($0.24 \pm 0.02$). SuStaIn GMM and KDE obtained $0.16 \pm 0.02$ and $0.13 \pm 0.02$, respectively. Random assignment yielded $0.00$. Subtype accuracy appeared insensitive to sample size, though increasing the healthy ratio impaired performance.

\textbf{Staging}. In staging tasks (Appendix~\ref{apd:res}, Fig.~\ref{fig:stage}), \textsc{bebms} assigned average stages of $0.16 \pm 0.03$ to control participants, while \textsc{bebms (blind)} achieved $0.62 \pm 0.09$. By comparison, SuStaIn KDE and GMM performed substantially worse, assigning average stages of $1.45 \pm 0.30$ and $3.03 \pm 0.24$, respectively. Importantly, SuStaIn’s staging accuracy degraded most severely when datasets contained fewer controls (i.e., smaller healthy ratios).

\textbf{Runtime}. \textsc{bebms} achieved better performance and was faster (Appendix~\ref{apd:res}, Fig.~\ref{fig:runtime}), with an average runtime across all datasets of $2.37 \pm 0.26$ minutes, followed by \textsc{bebms (blind)} ($4.04 \pm 0.43$ minutes). SuStaIn KDE ($6.57 \pm 0.58$ minutes) and SuStaIn GMM ($6.88 \pm 0.61$ minutes) took about twice as long.

\textbf{Subtype number estimation}. In estimating the optimal number of subtypes (Appendix~\ref{apd:res}, Fig.~\ref{fig:mae_cross}), \textsc{bebms} ($1.08 \pm 0.14$) performed marginally better than SuStaIn GMM ($1.16 \pm 0.20$), followed by \textsc{bebms (blind)} ($1.27 \pm 0.16$) and SuStaIn KDE ($1.92 \pm 0.23$) which had the same performance as random guessing ($1.92 \pm 0.33$). SuStaIn tends to overfit by identifying more subtypes than the ground truth, whereas both variants of \textsc{bebms} exhibit symmetric relative-error distributions centered at zero (See Fig.~\ref{fig:relative_error} in Appendix~\ref{apd:res}). Although \textsc{bebms} and SuStaIn GMM achieved comparable accuracy, \textsc{bebms} was faster (Appendix~\ref{apd:res}, Fig.~\ref{fig:runtime_cross}), requiring $40.11 \pm 15.28$ minutes on average compared to $63.68 \pm 15.11$ minutes for SuStaIn GMM.

\textbf{Stress-test experiments (Exp. 10--11).} We evaluated two settings with continuous event times under both EBM and Sigmoid generative frameworks. \textsc{bebms} consistently outperformed SuStaIn across all tasks. While performance on the ordering task degraded slightly, most algorithms showed improved subtyping and staging results compared to Exp.~1--9. See Figures~\ref{fig:extra_tau}--\ref{fig:extra_stage} in Appendix~\ref{apd:res}.

% the best-performing variants of each method achieved higher subtyping (Fig.~\ref{fig:extra_subtype}) and staging (Fig.~\ref{fig:extra_stage}) accuracy than in the standard experiments (Exp. 1–9). However, for ordering (Fig.~\ref{fig:extra_tau}), the performance of both methods degraded to the level of random guessing.

% \textbf{Stress-test experiments (Exp. 10-11)}. \textsc{bebms} outperformed SuStaIn across all tasks in the two stress-test experiments. Interestingly, the best-performing variants of each method achieved higher subtyping (Fig.~\ref{fig:extra_subtype}) and staging (Fig.~\ref{fig:extra_stage}) accuracy than in the standard experiments (Exp. 1–9). However, for ordering (Fig.~\ref{fig:extra_tau}), the performance of both methods degraded to the level of random guessing.

\subsection{ADNI}
\paragraph{Cross-validation and setup.}

We performed 5-fold cross-validation on the ADNI dataset to select the number of subtypes, testing values from 1 to 6 (the upper bound allowing us to assess potential overfitting). All other settings matched those used in the synthetic experiments. SuStaIn-GMM selected six subtypes, whereas \textsc{bebms} selected three (see explanations, Table~\ref{tab:cvic_comparison} and Figures~\ref{fig:cvic_bebms}--\ref{fig:cvic_sustain} in Appendix~\ref{apd:adni_res}). SuStaIn-KDE failed due to a singular matrix error. For both algorithms, we then ran ten replications with different random seeds and retained the solution with the highest data likelihood. For \textsc{bebms}, each run consisted of 20{,}000 MCMC iterations (200 burn-in, no thinning). Fig.~\ref{fig:adni_ordering_bebms} shows the inferred progression patterns for the three \textsc{bebms} subtypes based on the final 18{,}000 iterations. The trace plot indicates stable convergence (Fig.~\ref{fig:traceplot}, Appendix~\ref{apd:adni_res}).

In the results below, we present the inferred ordering, subtype assignments, and disease staging on ADNI. Biomarkers are grouped as CSF amyloid (A: $A\beta_{1-42}$), CSF tau (T: TAU, PTAU), cognition (C: ABETA, MMSE, ADAS13), and neurodegeneration (N: all remaining biomarkers; see Fig.~\ref{fig:adni_ordering_bebms}).

% We performed 5-fold cross-validation on the ADNI dataset to select the optimal number of subtypes, testing 1--6. The number of 6 was included to check for model overfitting. Other configurations matched those used in the synthetic experiments. SuStaIn GMM favored six subtypes, whereas \textsc{bebms} favored three (See explanations, Table \ref{tab:cvic_comparison}, Fig.~\ref{fig:cvic_bebms} and~\ref{fig:cvic_sustain} in Appendix~\ref{apd:adni_res}). SuStaIn KDE failed due to a singular matrix error. To obtain disease progressions, for both algorithms, we ran ten replicates with different random seeds and retained the run with the highest data likelihood. Note that in the above steps (ten replicates and the final run), for \textsc{bebms}, we ran 20,000 MCMC iterations with 200 burn in and no thinning. Fig.~\ref{fig:adni_ordering_bebms} is the progression of all three subtypes based on the last 18,000 iterations. The trace plot showed \textsc{bebms} had a clear sign of convergence (Fig.~\ref{fig:traceplot} in Appendix~\ref{apd:adni_res}). 

% In the following, we present the ordering, subtyping and staging results on ADNI. Biomarkers were grouped as CSF amyloid (A: $A\beta_{1-42}$), CSF tau (T: TAU and PTAU), and cognition (C:  ABETA, MMSE, and ADAS13), and neurodegeneration (N: all remaining biomarkers as in Fig. \ref{fig:adni_ordering_bebms}).

\paragraph{Ordering.}
\textsc{bebms} identified three subtypes:
\begin{enumerate}
    \item N $\to$ C $\to$ N $\to$ T $\to$ N $\to$ A\hfill (25, 3.4\%)
    \item N $\to$ T $\to$ A $\to$ C $\to$ N\hfill (493, 67.9\%) 
    \item N $\to$ C $\to$ A $\to$ T \hfill (208, 28.7\%)
\end{enumerate}

SuStaIn GMM identified six subtypes (See Fig.~\ref{fig:adni_ordering_sustain} in Appendix~\ref{apd:adni_res}):
\begin{enumerate}
    \item A $\to$ T $\to$ C $\to$ N $\to$ C $\to$ N \hfill (342, 47.1\%)
    \item A $\to$ N $\to$ C $\to$ N $\to$ C $\to$ T\hfill (124, 17.1\%)
    \item C $\to$ N $\to$ A $\to$ N $\to$ T \hfill (148, 20.4\%)
    \item N $\to$ C $\to$ A $\to$ T \hfill (54, 7.4\%)
    \item A $\to$ C $\to$ T $\to$ N \hfill (46, 6.3\%)
    \item T $\to$ N $\to$ C $\to$ N $\to$ C $\to$ N $\to$ A $\to$ N \hfill (12, 1.7\%)
\end{enumerate}

Table~\ref{tab:dx_composition_sustain} and~\ref{tab:dx_composition_bebms} in Appendix~\ref{apd:adni_res} have more detailed data of the distributions of subtypes over participants of different diagnoses. 

\paragraph{Staging.} 
\textsc{bebms} scored $1.10$ average stage, compared to $2.48$ for SuStaIn GMM for healthy patients. Fig.~\ref{fig:adni_staging_bebms} and~\ref{fig:adni_staging_sustain} in Appendix~\ref{apd:adni_res} have more details.

% \paragraph{Subtyping.}
% Subtyping does not have ground truth.

%% file: 6-discussion.tex
The ability of SuStaIn~\citep{young2018uncovering} to estimate subtypes of diseases from cross-sectional data has been a breakthrough for data-driven medical research. However, the robustness of its performance is unclear. In this paper, we presented a Bayesian variant of the EBM with subtypes (the \textsc{bebms}). Across a range of synthetic data experiments, we found that \textsc{bebms} performed better in all tasks: ordering, subtyping, staging, and estimating the number of subtypes, while being computationally more efficient. \textsc{bebms} can handle data with missing entries (whereas SuStaIn cannot) and shows great potential in scaling to high-dimensional data (See Appendix~\ref{apd:high_dimensional}). The blind variant of \textsc{bebms} has a slightly lower performance while also being computationally more demanding; domain expertise is needed to decide which version of our model to use, depending on the uncertainties in the diagnosis labels. 

\textsc{bebms}'s improved performance has important implications for clinical and research practice. Our model’s advantages can enhance clinical trial enrichment and patient stratification in precision medicine. Also, \textsc{bebms} remains robust across variation in the proportion of healthy participants, and the performance of all methods saturates at around 300. In many real-world settings, especially for new or rare diseases, recruiting large patient cohorts is difficult. Our results suggest that (1) cohorts larger than ~300 may not be necessary, and (2) when using \textsc{bebms}, increasing the proportion of healthy controls, which is typically far easier to recruit, has only a modest negative effect on performance.

The results on the real-world ADNI dataset show that \textsc{bebms} yields subtype patterns that align more closely with the current scientific consensus on Alzheimer’s disease progression than those identified by SuStaIn. Neuropathological evidence indicates three major AD subtypes~\citep{murray2011neuropathologically}. The most common subtype, Typical AD (TAD, $\sim$75\%), generally follows the ATNC sequence—amyloid (A) abnormality preceding tau (T), then neurodegeneration (N), and cognitive decline (C)~\citep{jack2024revised}, although only about one-third of patients strictly follow this ordering~\citep{mendes2025validating}. In contrast, Limbic-predominant AD (LPAD, $\sim$14\%) is characterized by early and severe involvement of the hippocampus and medial temporal cortex, with relatively limited neocortical involvement. Finally, Hippocampal-sparing AD (HSAD, $\sim$11\%) exhibits pronounced neocortical pathology—particularly in parietal and frontal association areas—while the hippocampus remains comparatively preserved.

Regarding the subtype patterns, \textsc{bebms} Subtype~2 (CSF-first, 67.9\%) corresponds to Typical AD (TAD), as CSF biomarkers become abnormal earliest; this aligns with SuStaIn Subtype~1 (47.1\%) and possibly Subtype~6 (1.7\%). \textsc{bebms} Subtype~3 (entorhinal/hippocampal-first, 28.7\%) corresponds to Limbic-predominant AD (LPAD), characterized by early medial temporal involvement; SuStaIn Subtypes~4 (7.4\%) and potentially Subtype~2 (17.1\%) reflect a similar pattern. Finally, \textsc{bebms} Subtype~1 (neocortical/cognitive-first with delayed hippocampal involvement, 3.4\%) corresponds to Hippocampal-sparing AD (HSAD), which matches SuStaIn Subtypes~3 (20.4\%) and~5 (6.3\%). 

Overall, \textsc{bebms} yields subtype proportions and progression patterns that more closely match the canonical findings reported in~\citet{murray2011neuropathologically}. However, \textsc{bebms} is not flawless: all subtypes show early entorhinal involvement, which may reflect mild overfitting or a strong entorhinal signal in ADNI. We caution against overinterpretation and believe our results should be treated as converging evidence to be assessed by domain experts, and not as definitive.

Our study has limitations. First, although \textsc{bebms} outperforms SuStaIn across tasks, its subtyping performance remains insufficient for clinical use. Second, our framework is limited to cross-sectional data; future work should explore subtype identification from longitudinal data, where models such as TEBM~\citep{wijeratne2023temporal} may be useful. Third, although \textsc{bebms} can handle missing values, we did not evaluate performance under varying degrees of missingness. Finally, our evaluation uses a single real-world dataset; further validation on additional datasets is needed to establish robustness.

%% file: appendix.tex
\section{ADNI Information}
\label{sec:adni}

% \subsection{ADNI Investigators}

A complete listing of ADNI
investigators can be found at:
\url{http://adni.loni.usc.edu/wp-content/uploads/how_to_apply/ADNI_Acknowledgement_List.pdf}

\section{Conjugate Prior Update for Distribution Parameters}
\label{apd:cp}

The updating rule:

$$W = \sum_{j=1}^J w_j, \bar{x} = \frac{1}{W}\sum_{j=1}^J w_j x_j$$

% $$\bar{x} = \frac{1}{W}\sum_{j=1}^J w_j x_j$$

$$S = \sum_{j=1}^J w_j (x_j - \bar{x})^2$$

$$m^\prime = \frac{n_0 m_0 + W \bar{x}}{n_0 + W}, n^\prime = n_0 + W$$

$$\nu^\prime = \nu_0 + W, s^\prime = \frac{1}{\nu^\prime}\left[S + \nu_0s_0^2 + \frac{n_0 W}{n^\prime}(\bar{x} - m_0)^2\right]$$

From these, the posterior mean of the mean is taken as:

$$\hat{\mu} = m^\prime,$$

\noindent and the posterior expectation of the variance is:

\begin{equation}
\hat{\sigma}^2 = \frac{\nu^\prime \cdot {s^\prime}^2/2}{\nu^\prime/2} = (s^\prime)^2
\end{equation}

% \begin{equation}
% \hat{\sigma}^2 \sim 
% \begin{cases}
% \frac{\nu^\prime {s^\prime}^2/2}{\nu^\prime/2 - 1}, & \nu^\prime > 2,\\
% \frac{\nu^\prime {s^\prime}^2}{\nu^\prime}, & \nu^\prime \le 2,
% \end{cases}
% \end{equation}

\noindent with posterior standard deviation $\hat{\sigma} = \sqrt{\hat{\sigma}^2}$. This initialization ensures that $\boldsymbol{\theta}, \boldsymbol{\phi}$ are informed by both clustering structure and prior uncertainty. 

Note that we knew the statistically correct calculation of $\hat{\sigma}^2$ should be:

\begin{equation}
\hat{\sigma}^2 =
\begin{cases}
\displaystyle \frac{\nu'(s')^2}{\nu' - 2}, & \nu' > 2,\\[8pt]
(s')^2, & \nu' \le 2.
\end{cases}
\end{equation}

\noindent but we decided to use $\hat{\sigma}^2 = (s^\prime)^2$ because it led to better empirical results.

\newpage 
\section{\textsc{BEBMS} Inference Algorithm}
\label{apd:inference_algo}

\begin{algorithm2e}[htbp]
\label{algo:mcmc}
\caption{\textsc{bebms} Metropolis--Hastings Sampler}
\KwIn{Data $\boldsymbol{X} \in \mathbb{R}^{J \times N}$, Number of subtypes $T$}
\KwOut{Samples
$\{\boldsymbol{S}^{(i)}, \boldsymbol{\theta}^{(i)}, \boldsymbol{\phi}^{(i)}, 
  \boldsymbol{\pi}_t^{(i)}, \boldsymbol{\pi}_{k\mid t}^{(i)}, 
  P_{\text{stage}}^{(i)}, P_{\text{subtype}}^{(i)}, \ell^{(i)}\}_{i=1}^M$}

\textbf{Init:}  
$\boldsymbol{S}^{(0)}, \boldsymbol{\theta}^{(0)}, \boldsymbol{\phi}^{(0)}, 
 \boldsymbol{\alpha}_t, \boldsymbol{\alpha}_{k|t}, 
 \boldsymbol{\pi}_t^{(0)}, \boldsymbol{\pi}_{k\mid t}^{(0)}$\;  
Compute initial posteriors $P_{\text{stage}}^{(0)}, P_{\text{subtype}}^{(0)}$\;  
Compute total log-likelihood $\ell^{(0)}$\;

\For{$i=1$ \KwTo $M$}{
  \tcp{Step 1: Propose new orderings}
  Propose $\boldsymbol{S^\prime}$ from $q(\boldsymbol{S^\prime} \mid \boldsymbol{S}^{(i-1)})$\;
  Compute intermediate posteriors 
  $\tilde{P}_{\text{stage}}, \tilde{P}_{\text{subtype}}$ 
  using $\boldsymbol{S^\prime}$ and $(\boldsymbol{\theta}^{(i-1)}, \boldsymbol{\phi}^{(i-1)})$\;
  Update $(\boldsymbol{\theta}^\prime, \boldsymbol{\phi}^\prime)$ using $\boldsymbol{S^\prime},\tilde{P}_{\text{stage}}, \tilde{P}_{\text{subtype}}$\;
  
  \tcp{Step 2: Compute likelihood}
  Recompute $P_{\text{stage}}^\prime, P_{\text{subtype}}^\prime$ and $\ell^\prime$\;
  
  \tcp{Step 3: Acceptance probability}
  $\alpha \gets \min\!\left(1, e^{\ell^\prime - \ell^{(i-1)}}\right)$\;
  $U \sim \text{Uniform}(0, 1)$\;
  \tcp{Step 4: Accept/Reject}
  \If{$U<\alpha$}{
    $S^{(i)}\gets \boldsymbol{S^\prime}$\;
    $\boldsymbol{\theta}^{(i)}\gets\boldsymbol{\theta}^\prime$, 
    $\boldsymbol{\phi}^{(i)}\gets\boldsymbol{\phi}^\prime$\;
    $P_{\text{stage}}^{(i)}\gets P_{\text{stage}}^\prime$, 
    $P_{\text{subtype}}^{(i)}\gets P_{\text{subtype}}^\prime$\;
    $\ell^{(i)}\gets \ell^\prime$\;
    % \tcp{Step 5: Update priors}
    $\boldsymbol{\pi}_t^{(i)} \sim \text{Dir}(\boldsymbol{\alpha}_t + \text{counts})$\;
    $\boldsymbol{\pi}_{k\mid t}^{(i)} \sim \text{Dir}(\boldsymbol{\alpha}_{k\mid t} + \text{counts})$\;
  }
}
\end{algorithm2e}

At each MCMC iteration, we propose a new event sequence $\boldsymbol{S^\prime}$ by randomly selecting two subtypes. For each of these two subtypes, we then randomly select two biomarkers and swap their positions within the subtype's ordering. If there is only one subtype, we randomly select two biomarkers from it and swap their positions. This defines the proposal distribution $q(\boldsymbol{S^\prime} \mid \boldsymbol{S}^{(i-1)})$. 

\clearpage
\section{ADNI Select Biomarker Glossary}
\label{apd:glossary}

\begin{table*}[ht]
\centering
\caption{Glossary of ADNI Biomarkers with Source, Units, and Interpretation}
\label{tab:biomarker_glossary}
\scriptsize
\resizebox{\textwidth}{!}{%
\begin{tabular}{
  l
  >{\raggedright\arraybackslash}p{3.8cm}
  >{\raggedright\arraybackslash}p{2.8cm}
  >{\raggedright\arraybackslash}p{2.2cm}
  >{\raggedright\arraybackslash}p{5.2cm}
}
\toprule
\textbf{Abbrev.} & \textbf{Full Name} & \textbf{Source Modality} & \textbf{Unit / Scale} & \textbf{Higher Values Indicate} \\
\midrule
MMSE        & Mini-Mental State Examination & Cognitive test  & Score (0--30)  & Less pathology (better global cognition) \\
ADAS13      & Alzheimer’s Disease Assessment Scale -- Cognitive Subscale (13-item) & Cognitive test  & Score (0--85)  & More pathology (worse cognition) \\
RAVLT-immediate & Rey Auditory Verbal Learning Test -- Immediate Recall  & Cognitive test  & Score (0--75)  & Less pathology (better memory encoding) \\
ABETA       & Amyloid Beta (A$\beta_{1-42}$) & CSF            & pg/mL          & Less pathology (less amyloid deposition) \\
TAU         & Total Tau                      & CSF            & pg/mL          & More pathology (axonal degeneration / neuronal injury) \\
PTAU        & Phosphorylated Tau (p-Tau$_{181}$) & CSF         & pg/mL          & More pathology (neurofibrillary tangle burden) \\
Ventricles  & Ventricular Volume (ICV-normalized) & Structural MRI & Fraction of ICV & More pathology (greater brain atrophy) \\
WholeBrain  & Whole Brain Volume (ICV-normalized) & Structural MRI & Fraction of ICV & Less pathology (greater structural integrity) \\
Hippocampus & Hippocampal Volume (ICV-normalized) & Structural MRI & Fraction of ICV & Less pathology (greater structural integrity) \\
Entorhinal  & Entorhinal Cortex Volume (ICV-normalized) & Structural MRI & Fraction of ICV & Less pathology (greater structural integrity) \\
Fusiform    & Fusiform Gyrus Volume (ICV-normalized) & Structural MRI & Fraction of ICV & Less pathology (greater structural integrity) \\
MidTemp     & Middle Temporal Gyrus Volume (ICV-normalized) & Structural MRI & Fraction of ICV & Less pathology (greater structural integrity) \\
\bottomrule
\end{tabular}}
\end{table*}

\clearpage 
\section{Biomarker Parameters and Non-Normal Distribution Parameter Details}
\label{apd:biomarker_params_and_irr}

\begin{table}[ht]
\centering
\caption{Biomarker Parameterization and Irregular Sampling Distributions}
\label{tab:biomarker_params_and_irr}
\scriptsize
\begin{tabular}{lccccp{7.5cm}}
\toprule
\textbf{Biomarker} & $\theta_{\text{mean}}$ & $\theta_{\text{std}}$ & $\phi_{\text{mean}}$ & $\phi_{\text{std}}$ & \textbf{Irregular Distribution (Per Implementation)} \\
\midrule

MMSE & 25.31 & 2.38 & 29.17 & 0.81 &
Triangular$(\mu-2\sigma, \mu-1.5\sigma, \mu)$; 
$\mathcal{N}(\mu+\sigma, (0.3\sigma)^2)$;
Exp$(0.7\sigma)+(\mu-0.5\sigma)$ (equal mixture). \\

ADAS13 & 21.79 & 9.51 & 9.32 & 3.91 &
Same mixture structure as MMSE (triangular + Gaussian + exponential). \\

RAVLT\_immediate & 27.50 & 7.93 & 45.39 & 9.36 &
Same mixture structure as MMSE (triangular + Gaussian + exponential). \\
\midrule

ABETA & 661.23 & 195.29 & 1331.37 & 214.57 &
Pareto(1.5)$\cdot\sigma+(\mu-2\sigma)$; 
$\mathcal{U}(\mu-1.5\sigma,\mu+1.5\sigma)$; 
Logistic$(\mu,\sigma)$ (equal mixture). \\

TAU & 385.84 & 138.95 & 208.11 & 58.84 &
Same mixture structure as ABETA. \\

PTAU & 37.21 & 15.09 & 17.88 & 5.13 &
Same mixture structure as ABETA. \\
\midrule

VentricleNorm & 0.0359 & 0.0128 & 0.0198 & 0.0069 &
Beta(0.5,0.5)$\cdot4\sigma+(\mu-2\sigma)$; 
Exp$(0.4\sigma)$ with $\pm$ sign; 
$\mathcal{N}(\mu,(0.5\sigma)^2)$ + {0, $2\sigma$} spike. \\

HippocampusNorm & 0.00390 & 0.00065 & 0.00511 & 0.00059 &
Same mixture structure as VentricleNorm. \\
\midrule

WholeBrainNorm & 0.6311 & 0.0346 & 0.6949 & 0.0389 &
Gamma$(2, 0.5\sigma)+(\mu-\sigma)$; 
Weibull$(1.0)\cdot\sigma+(\mu-\sigma)$; 
$\mathcal{N}(\mu,(0.5\sigma)^2)\pm\sigma$. \\

EntorhinalNorm & 0.00217 & 0.00050 & 0.00253 & 0.00038 &
Same mixture structure as WholeBrainNorm. \\
\midrule

FusiformNorm & 0.01116 & 0.00167 & 0.01186 & 0.00140 &
Standard Cauchy$(\mu,\sigma)$ + $\mathcal{N}(0,(0.2\sigma)^2)$, clipped to $[\mu-4\sigma,\mu+4\sigma]$. \\
\midrule

MidTempNorm & 0.01241 & 0.00179 & 0.01344 & 0.00140 &
10\% $\mathcal{N}(\mu,0.2\sigma)$ spike + 90\% Logistic$(\mu+\sigma,2\sigma)$. \\
\bottomrule
\end{tabular}

\vspace{1mm}
\raggedright
\small
\textbf{Implementation Notes:} 

\begin{itemize}
    \item $\mu$ \& $\sigma$ use $\theta$ parameters for affected (pathological) and $\phi$ for nonaffected (intact). 
    \item For non-normal components, \textbf{After sampling, all values are perturbed by additional noise} $\mathcal{N}(0,\ (0.2\sigma)^2)$ \textbf{and clipped to} $[\mu - 5\sigma,\ \mu + 5\sigma]$.
\end{itemize}
\end{table}

\newpage \section{Experiment Specifications}
\label{apd:experiment_specifications}

\begin{enumerate}
    \item Ordinal $\boldsymbol{S}$ \& Ordinal $k_j$ with bell-shape Dirichlet priors \& $x_{j,n}$ sampled from normal distributions according to the EBM model.

    \item Same as Experiment 1, but non-normal distributions for $x_{j,n}$. 

    \item Same as Experiment 1, but with uniform distributions for ordinal $k_j$.

    \item Same as Experiment 2, but with uniform distributions for ordinal $k_j$.

    \item Ordinal $\boldsymbol{S}$ \& Continuous $k_j$ with uniform distributions \&  $x_{j,n}$ sampled from normal distributions according to the EBM model.

    \item Same as Experiment 5, but with non-normal distributions for $x_{j,n}$. 

    \item Same as Experiment 6, but with a scaled Beta distribution ($\lambda = N, \alpha = 5, \beta = 2$) for $k_j$.

    \item Ordinal $\boldsymbol{S}$ \& Continuous $k_j$ with uniform distributions \&  $x_{j,n}$ generated using the Sigmoid model. 

    \item Same as Experiment 8, but with a scaled Beta distribution ($\lambda = N, \alpha = 5, \beta = 2$) for $k_j$. 

    \item Continuous event times ($\text{Beta}(2,2) \times N$) \& scaled Beta distribution ($\lambda = N, \alpha = 5, \beta = 2$) for $k_j$ \& $x_{j,n}$ sampled from normal distributions according to the EBM model.

    \item Same as Experiment 10, but $x_{j,n}$ was generated using the Sigmoid model. 

\end{enumerate} 

\clearpage 
\section{Theoretical and Empirical Biomarker Distributions}
\label{apd:all-dist}
\begin{figure}[htbp]
   \centering 
   \includegraphics[scale=0.3]{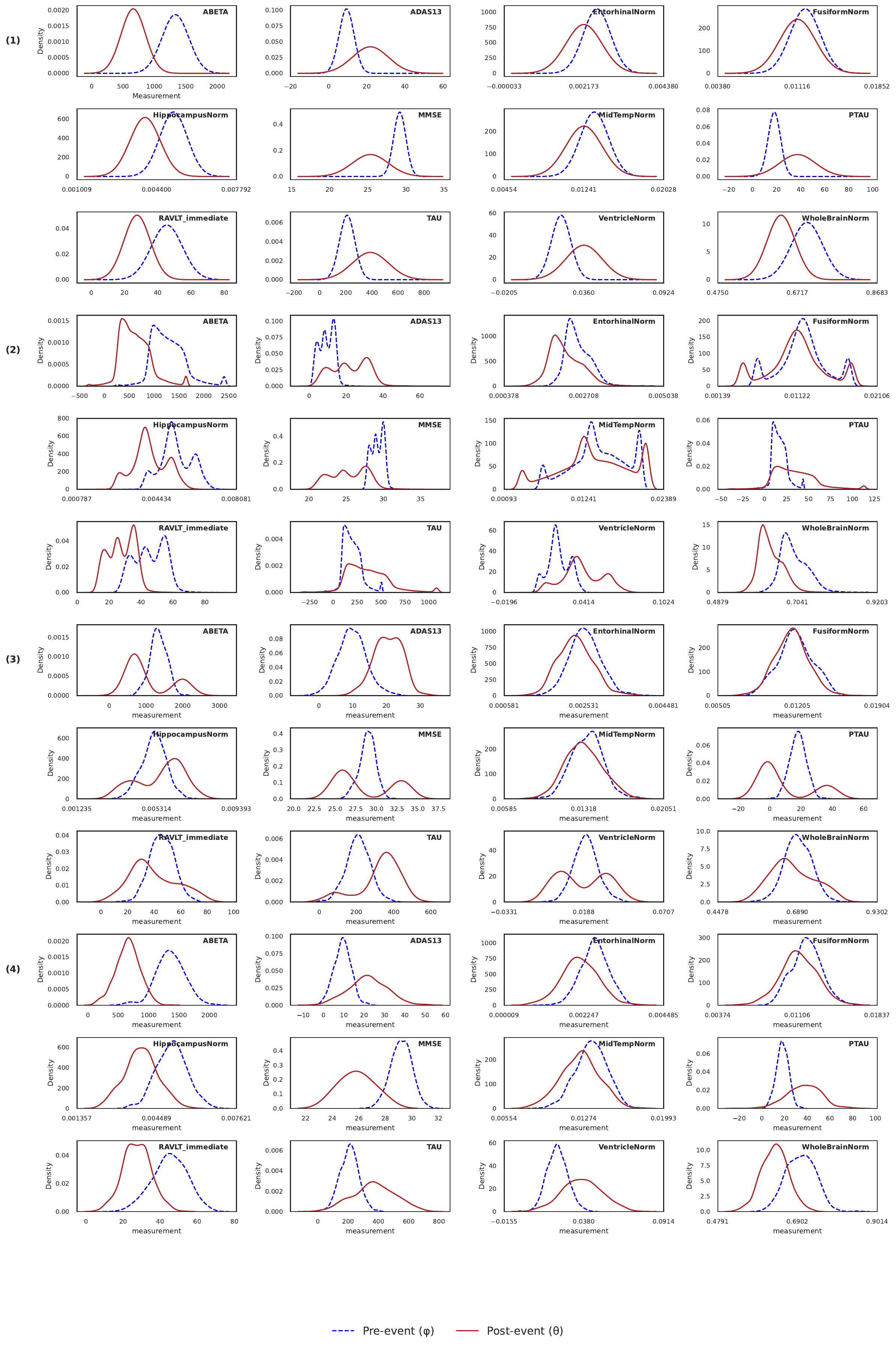} 
   \caption{(1) Theoretical normal distributions; (2) Theoretical non-normal distributions; (3) Empirical distributions in one synthetic dataset of Exp. 9; (4) Empirical distributions in one synthetic dataset of Exp. 1.}
    \label{fig:all-dist}
\end{figure}

\clearpage
\section{Kendall's $W$}
\label{apd:kendalls_w}

We used Kendall's $W$ to measure the similarity among disease progression of different subtypes. $W$ ranges from 0 to 1, with 0 indicating no similarity at all and 1 indicating exactly the same. We aim for the full range of $[0,1]$ in synthetic experiments because we want to make sure the performances of all algorithms are not dependent on the similarity among subtypes. 

\begin{figure}[htbp]
   \centering 
   \includegraphics[scale=0.7]{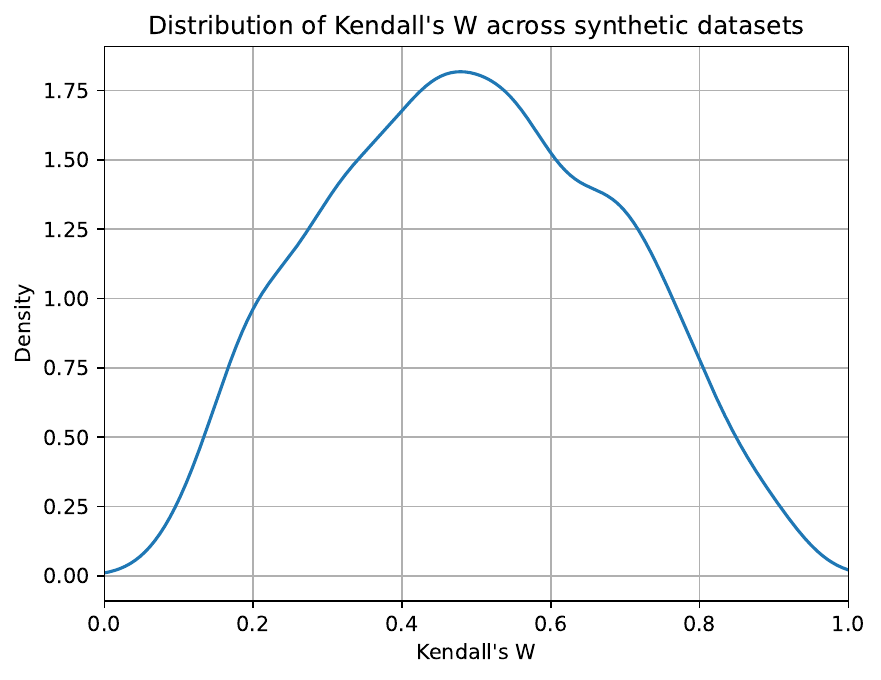} 
   \caption{Distribution of Kendall's $W$ across all 1,318 synthetic datasets (Exp. 1-11).}
    \label{fig:kendalls_w}
\end{figure}

\clearpage
\section{Synthetic Experiment Results}\label{apd:res}

\begin{figure*}[htbp]
 % Caption and label go in the first argument and the figure contents
 % go in the second argument
\floatconts
  {fig:subtype}
  {\vspace{-1cm}\caption{Subtyping results, measured with Adjusted Rand Index (ARI). Larger values reflect better performance. Subtyping is hard for both \textsc{bebms} and SuStaIn. 
  }}
  {\includegraphics[width=1.05\linewidth]{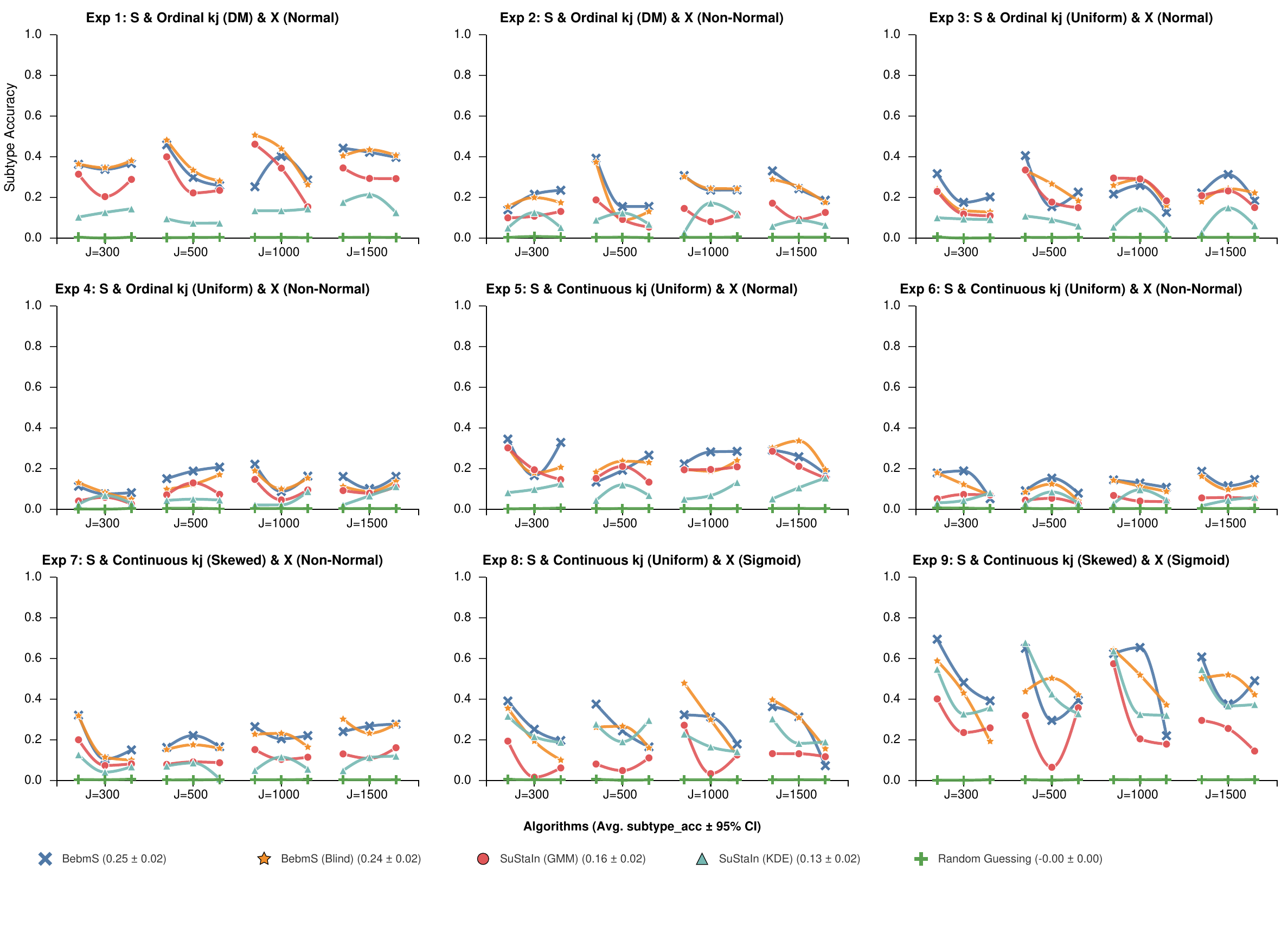}}
\end{figure*}

\begin{figure*}[htbp]
 % Caption and label go in the first argument and the figure contents
 % go in the second argument
\floatconts
  {fig:stage}
  {\vspace{-1cm}\caption{Staging results. \textsc{bebms} assigned lower disease stages to control participants than SuStaIn. 0 is the ground truth. 
  }}
  {\includegraphics[width=1.05\linewidth]{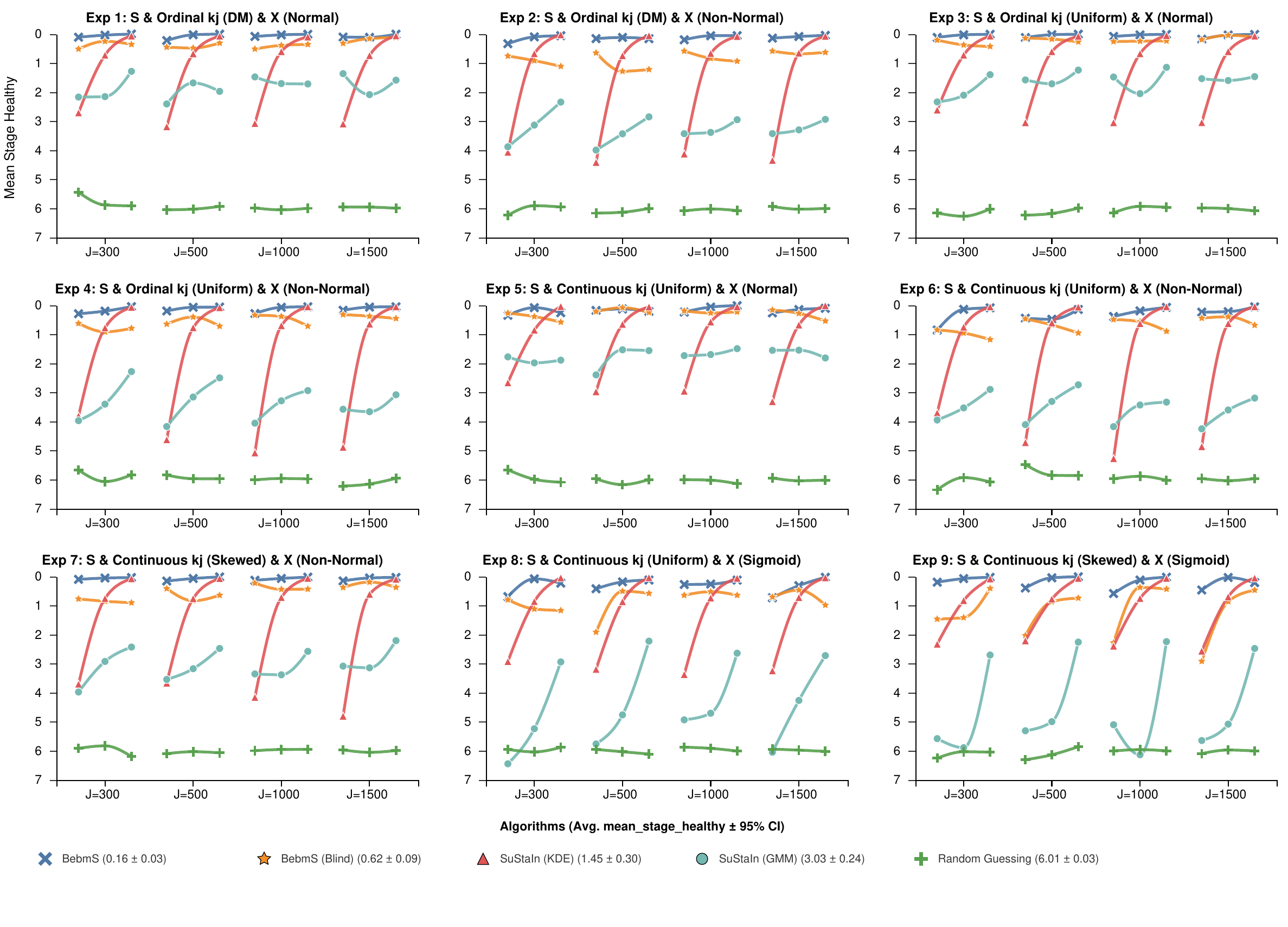}}
\end{figure*}

\begin{figure*}[htbp]
 % Caption and label go in the first argument and the figure contents
 % go in the second argument
\floatconts
  {fig:runtime}
  {\vspace{-1cm}\caption{Runtime analysis. \textsc{bebms} is much faster than SuStaIn. 
  }}
  {\includegraphics[width=1.05\linewidth]{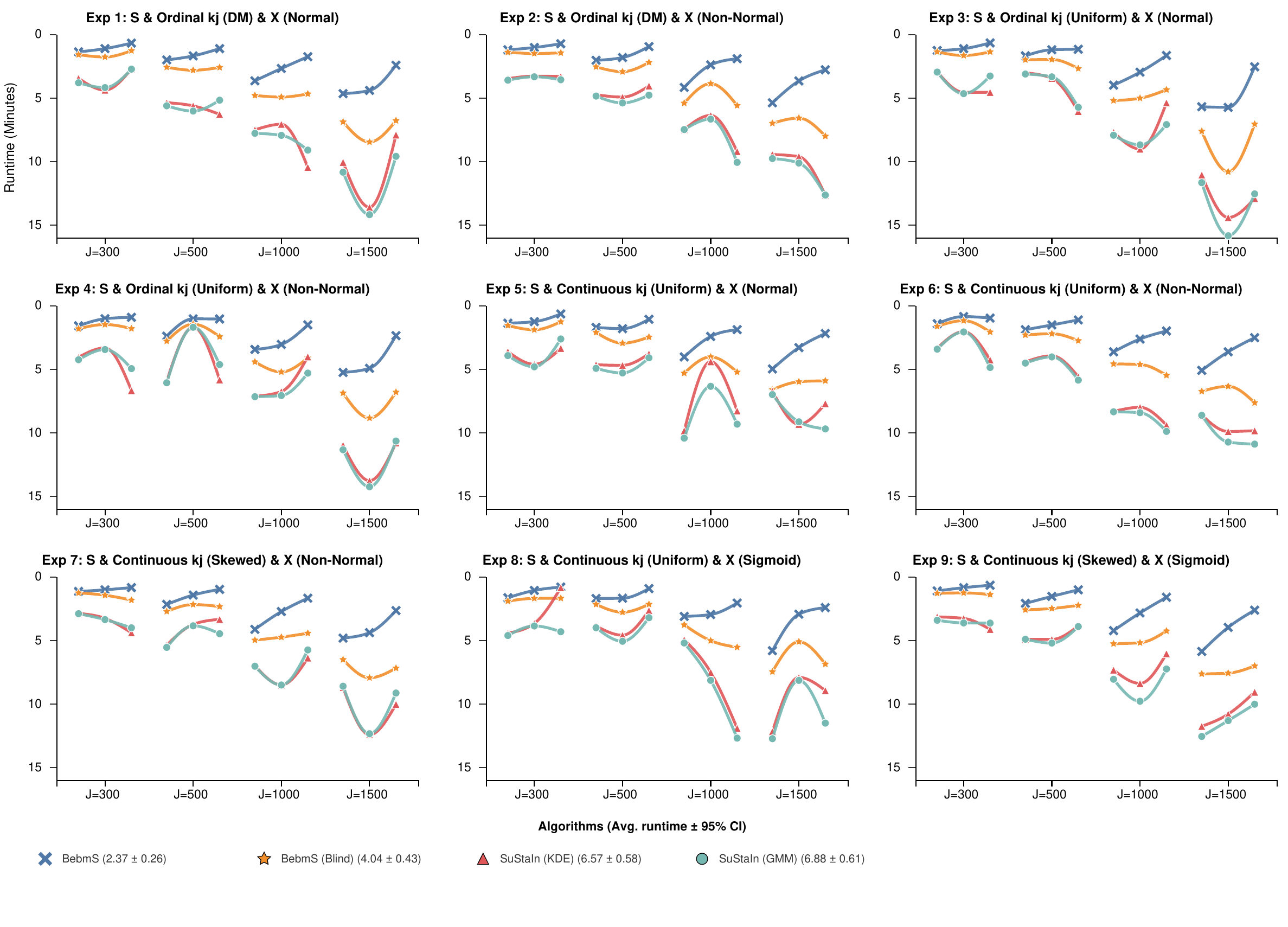}}
\end{figure*}

\begin{figure}[htbp]
 % Caption and label go in the first argument and the figure contents
 % go in the second argument
\floatconts
  {fig:mae_cross}
  {\vspace{-1cm}\caption{Mean Absolute Error (MAE) for the estimation of ground-truth number of subtypes. Smaller values reflect better performance. \textsc{bebms}'s performance is similar to that of SuStaIn GMM. 
  }}
  {\includegraphics[width=0.5\linewidth]{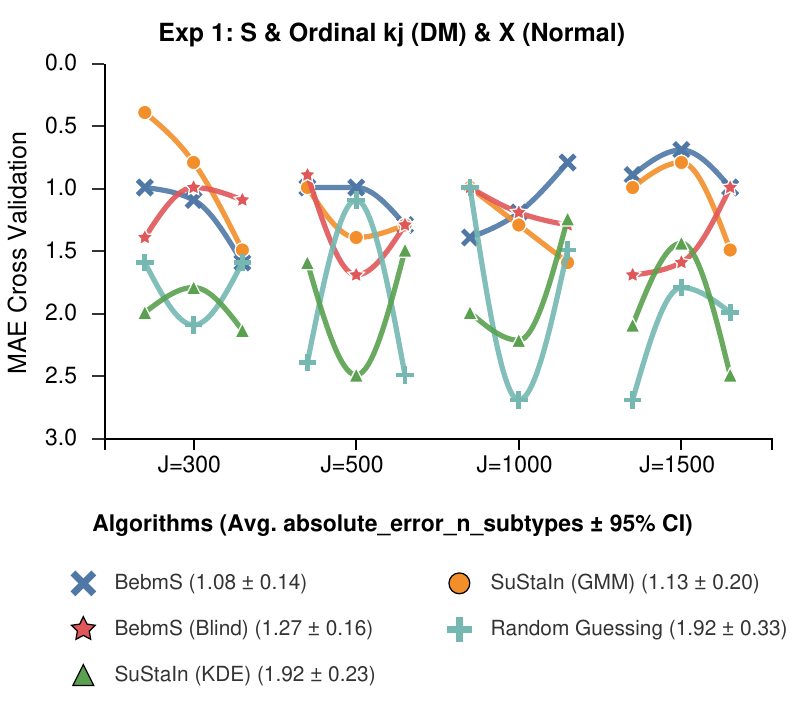}}
\end{figure}

\begin{figure}[htbp]
 % Caption and label go in the first argument and the figure contents
 % go in the second argument
\floatconts
  {fig:runtime_cross}
  {\vspace{-1cm}\caption{Runtime analysis for the estimation of ground-truth number of subtypes. \textsc{bebms} achieved similar accuracy to SuStaIn GMM but with a faster speed. 
  }}
  {\includegraphics[width=0.5\linewidth]{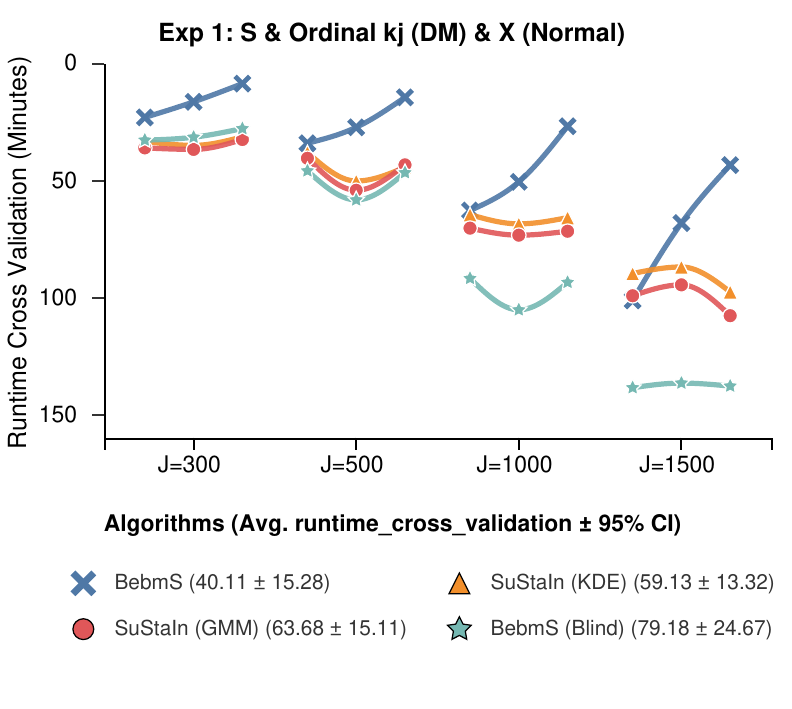}}
\end{figure}

\begin{figure}[htbp]
 % Caption and label go in the first argument and the figure contents
 % go in the second argument
\floatconts
  {fig:relative_error}
  {\vspace{-1cm}\caption{Relative error distributions for estimating the optimal subtype count. The \textsc{bebms} variants produce symmetric, zero-centered errors, while SuStaIn consistently overestimates (overfits) the number of subtypes.
  }}
  {\includegraphics[width=0.7\linewidth]{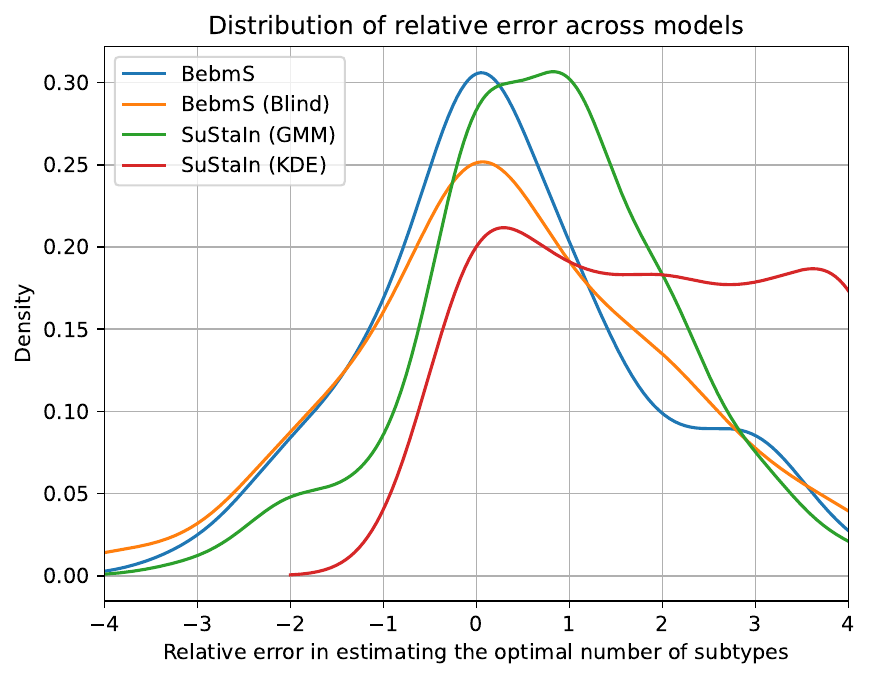}}
\end{figure}

\begin{figure}[htbp]
 % Caption and label go in the first argument and the figure contents
 % go in the second argument
\floatconts
  {fig:extra_tau}
  {\vspace{-0.5cm}\caption{Ordering results for stress-test experiments (Kendall's tau distance).
  }}
  {\includegraphics[width=1\linewidth]{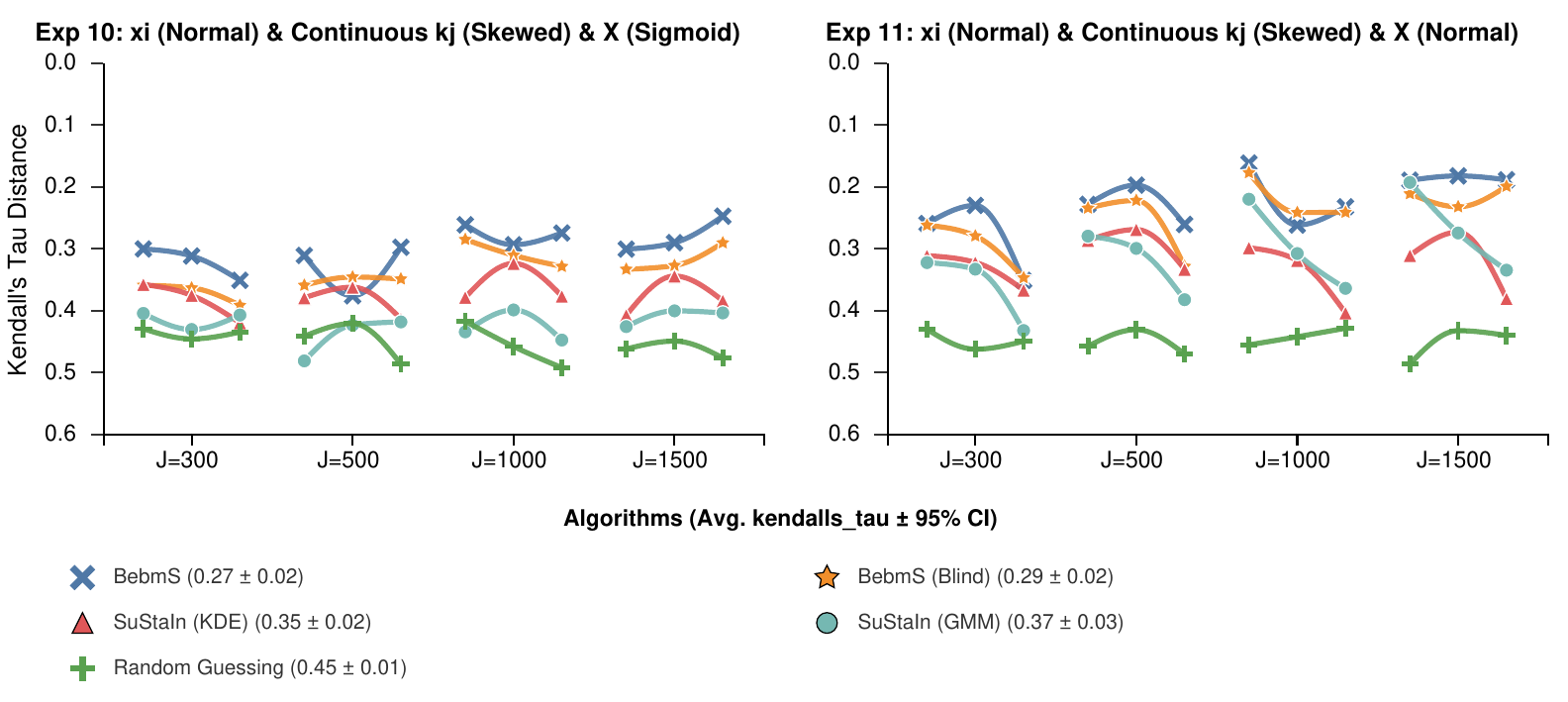}}
\end{figure}

\begin{figure}[htbp]
 % Caption and label go in the first argument and the figure contents
 % go in the second argument
\floatconts
  {fig:extra_subtype}
  {\vspace{-0.5cm}\caption{Subtype assignment results for stress-test experiments. .
  }}
  {\includegraphics[width=1\linewidth]{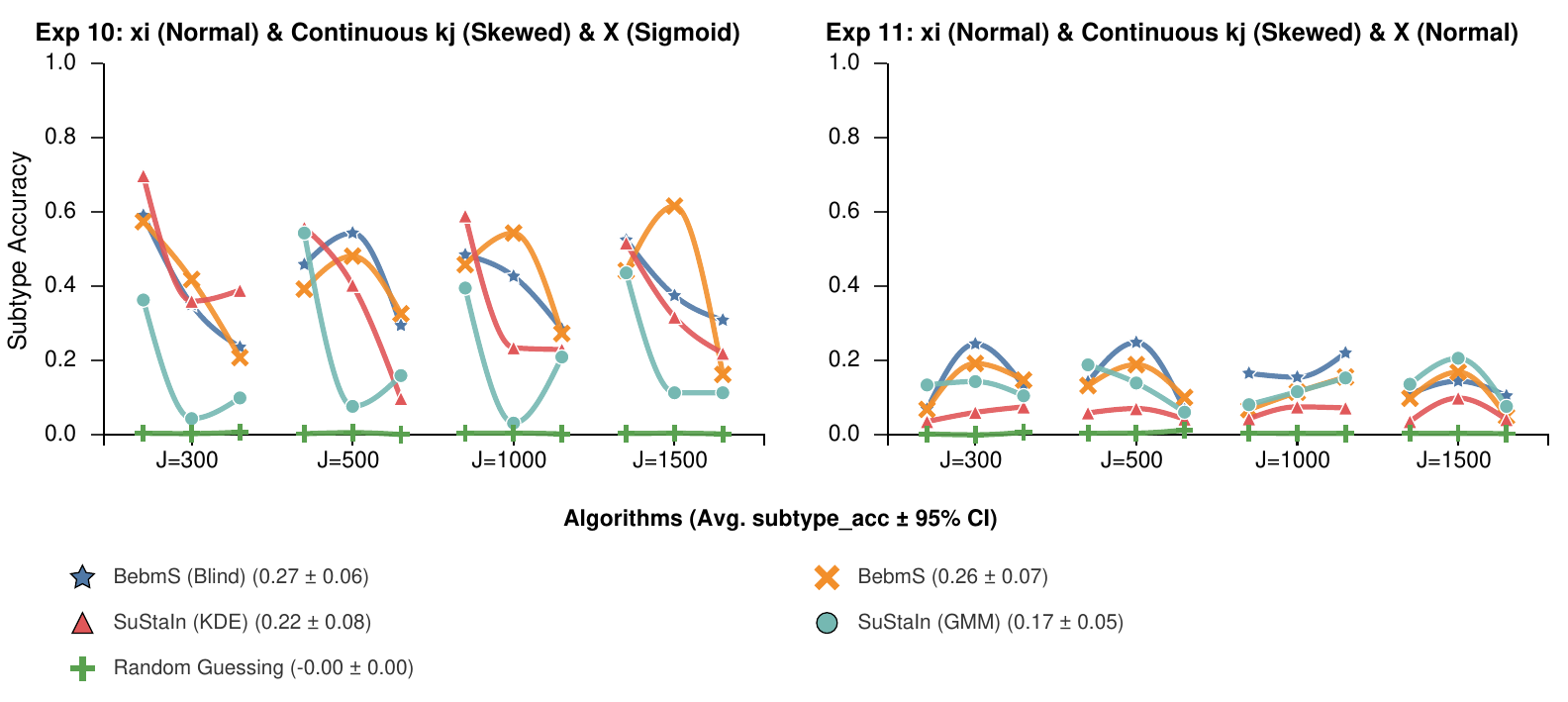}}
\end{figure}

\begin{figure}[htbp]
 % Caption and label go in the first argument and the figure contents
 % go in the second argument
\floatconts
  {fig:extra_stage}
  {\vspace{-0.5cm}\caption{Staging on stress-test experiments.
  }}
  {\includegraphics[width=1\linewidth]{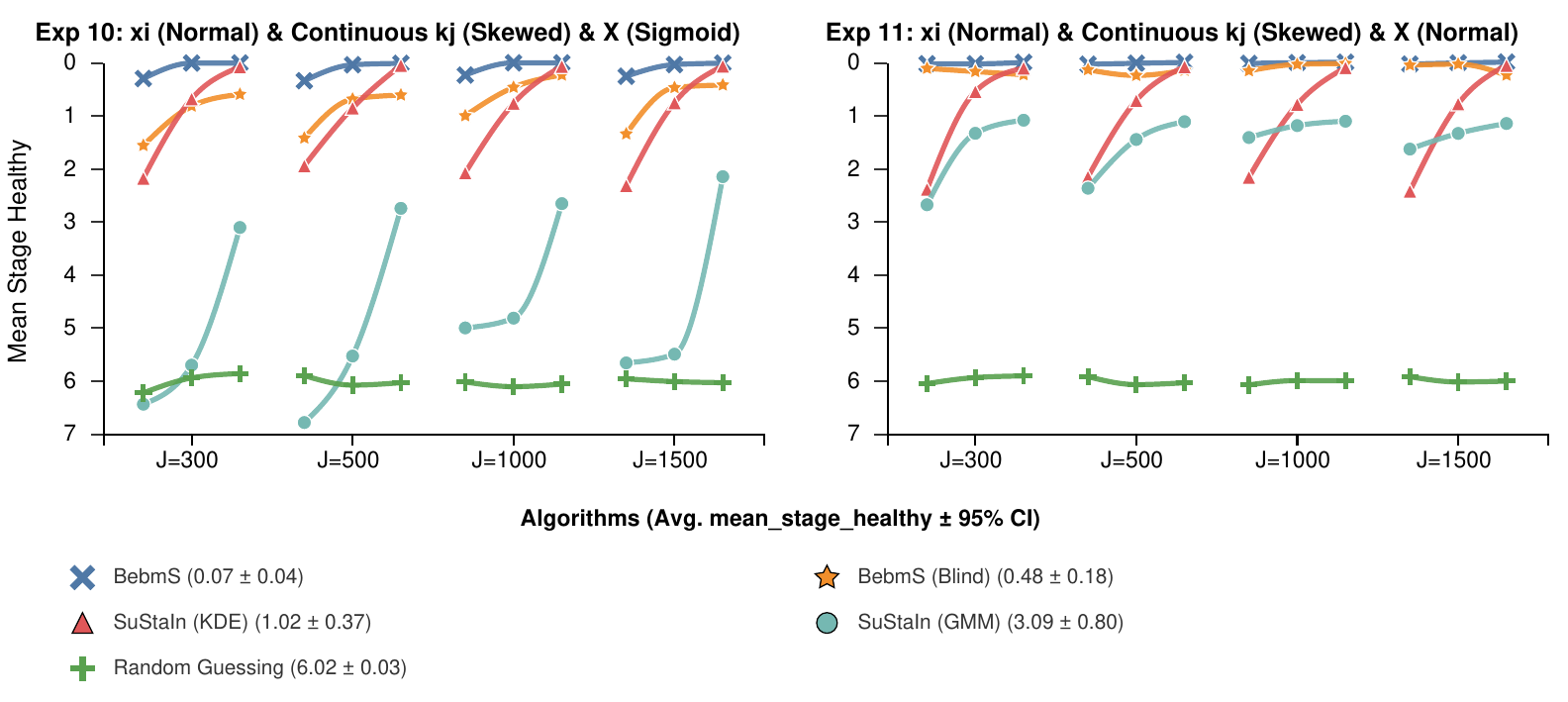}}
\end{figure}

\clearpage \section{ADNI Results}\label{apd:adni_res}

The results vary because of randomness. We tested \textsc{bebms} and SuStaIn GMM several times. \textsc{bebms} favored 3 or 4 as the optimal number of subtypes, and SuStaIn favored 5 or 6. We picked 6 for SuStaIn because that was our last attempt. For \textsc{bebms}, when the number of subtypes was 4, one of the resulting subtypes only had 6 participants, a clear indication of overfitting. Therefore, we chose 3.

\begin{figure}[ht]
\centering

\begin{minipage}{0.46\textwidth}
  \centering
  \includegraphics[width=\textwidth]{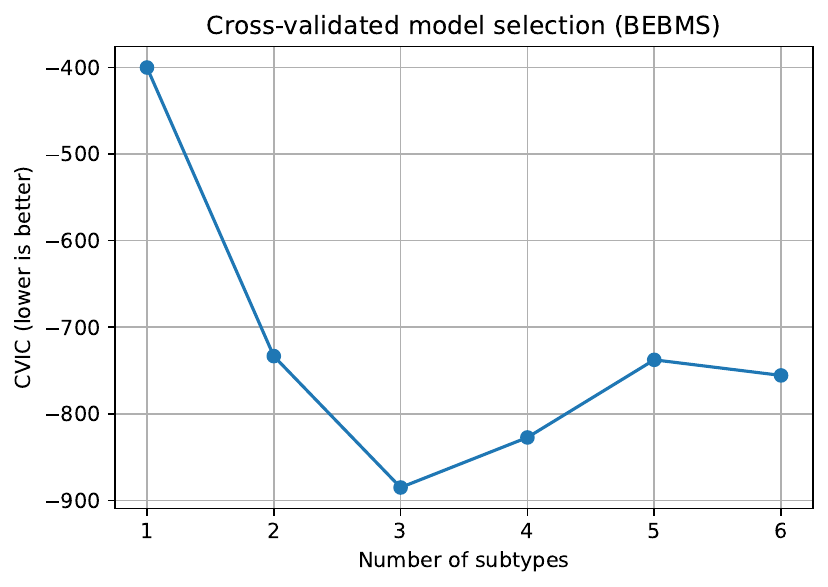}
  \caption{Cross-validation on ADNI using \textsc{bebms}}
  \label{fig:cvic_bebms}
\end{minipage}
\hfill
\begin{minipage}{0.46\textwidth}
  \centering
  \includegraphics[width=\textwidth]{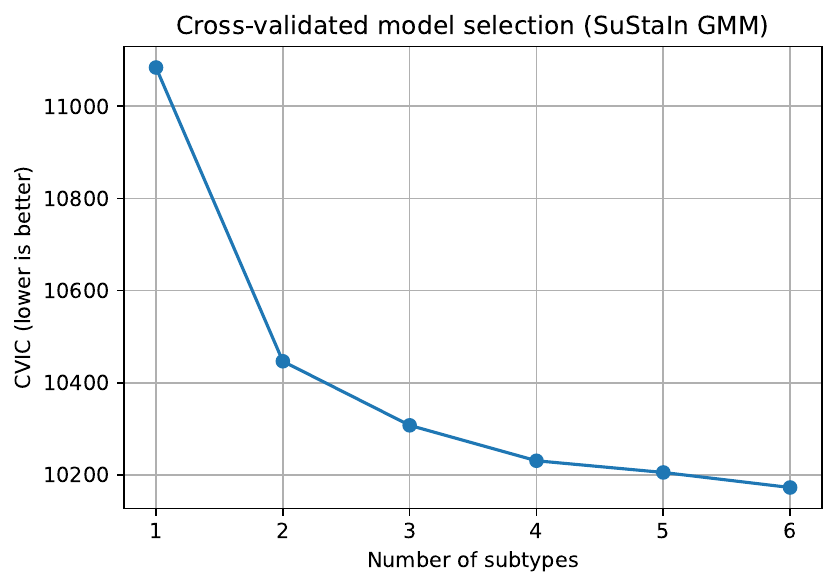}
  \caption{Cross-validation on ADNI using SuStaIn (GMM)}
  \label{fig:cvic_sustain}
\end{minipage}
\end{figure}

\begin{table}[ht]
\centering
\caption{Comparison of CVIC across number of subtypes for \textsc{bebms} and SuStaIn (GMM).}
\vspace{0.2cm}
\begin{tabular}{c c c}
\hline
\textbf{Number of Subtypes} & \textbf{BEBMS CVIC} & \textbf{SuStaIn (GMM) CVIC} \\
\hline
1 & $-400.11$  & $11084.15$ \\
2 & $-733.31$  & $10446.74$ \\
3 & $\mathbf{-885.00}$ & $10307.71$ \\
4 & $-827.21$  & $10230.84$ \\
5 & $-737.59$  & $10205.47$ \\
6 & $-755.60$  & $\mathbf{10172.81}$ \\
\hline
\end{tabular}
\par\vspace{0.2cm}
\small\textit{Note.} CVIC = Cross-Validation Information Criterion. Lower values indicate better model fit.
\label{tab:cvic_comparison}
\end{table}

\begin{figure}[htbp]
 % Caption and label go in the first argument and the figure contents
 % go in the second argument
\floatconts
  {fig:traceplot}
  {\vspace{-0.5cm}\caption{The trace plot of running \textsc{bebms} on ADNI. The starting point is iteration 40. Qualitatively, this suggests good convergence.
  }}
  {\includegraphics[width=0.9\linewidth]{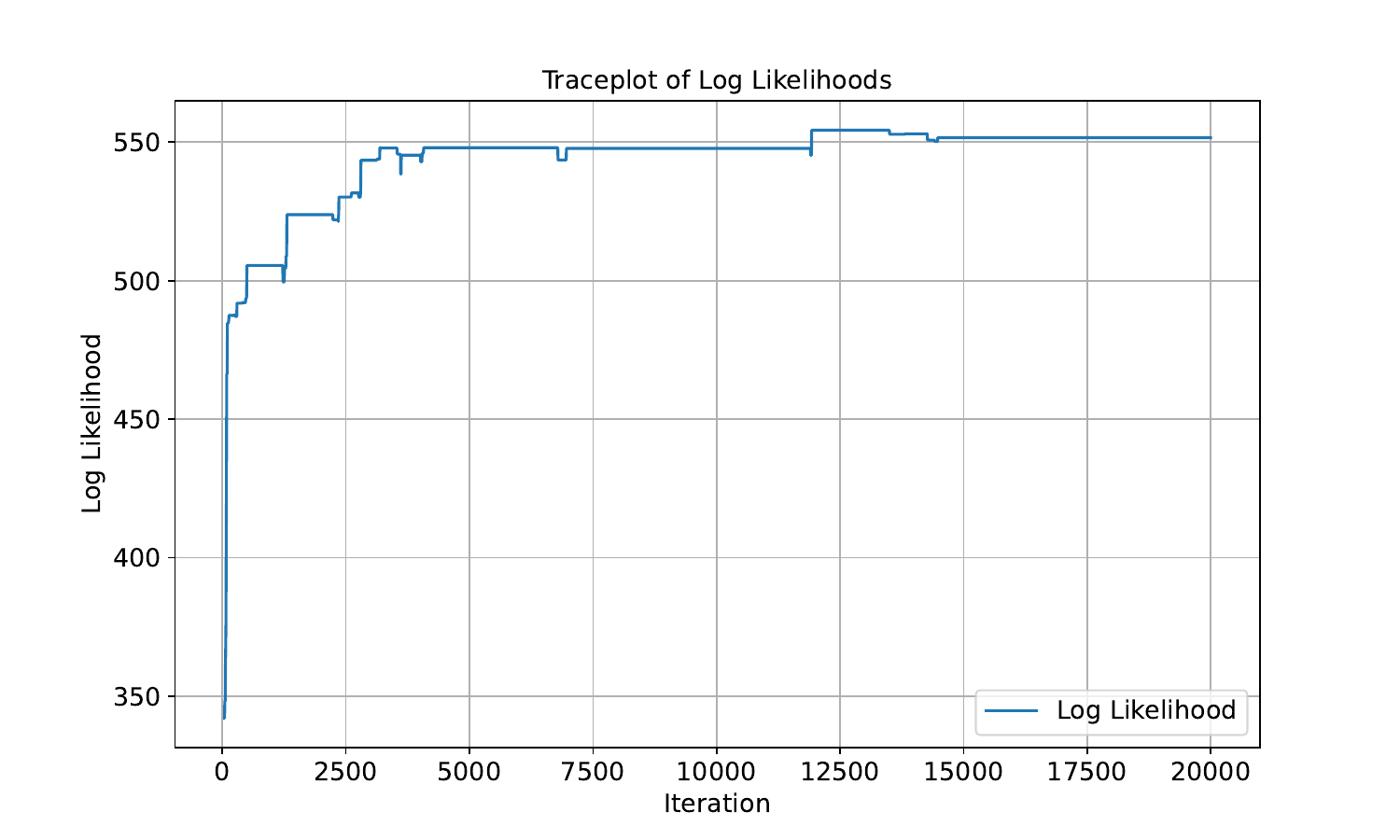}}
\end{figure}

\begin{table}[ht]
\centering
\caption{Diagnostic composition (proportions) across subtypes (SuStaIn GMM)}
\vspace{0.2cm}
\label{tab:dx_composition_sustain}
\begin{tabular}{lccccc}
\hline
\textbf{Subtype} & \textbf{Total} & \textbf{AD} & \textbf{CN} & \textbf{EMCI} & \textbf{LMCI} \\
\hline
1 & 342 & 0.31 & 0.16 & 0.21 & 0.32 \\
2 & 124 & 0.03 & 0.30 & 0.32 & 0.35 \\
3 & 148 & 0.27 & 0.14 & 0.25 & 0.34 \\
4 & 54  & 0.00 & 0.35 & 0.28 & 0.37 \\
5 & 46  & 0.02 & 0.39 & 0.28 & 0.30 \\
6 & 12  & 0.17 & 0.42 & 0.33 & 0.08 \\
\hline
\end{tabular}
\end{table}

\begin{table}[ht]
\centering
\caption{Diagnostic composition (proportions) across subtypes (\textsc{bebms})}
\vspace{0.2cm}
\label{tab:dx_composition_bebms}
\begin{tabular}{lccccc}
\hline
\textbf{Subtype} & \textbf{Total} & \textbf{AD} & \textbf{CN} & \textbf{EMCI} & \textbf{LMCI} \\
\hline
1 & 25  & 0.28 & 0.04 & 0.28 & 0.40 \\
2 & 493 & 0.17 & 0.26 & 0.30 & 0.28 \\
3 & 208 & 0.30 & 0.13 & 0.13 & 0.43 \\
\hline
\end{tabular}
\end{table}

\begin{figure}[htbp]
 % Caption and label go in the first argument and the figure contents
 % go in the second argument
\floatconts
  {fig:adni_staging_bebms}
  {\vspace{-0.5cm}\caption{\textsc{bebms} ADNI staging and subtyping
  }}
  {\includegraphics[width=0.8\linewidth]{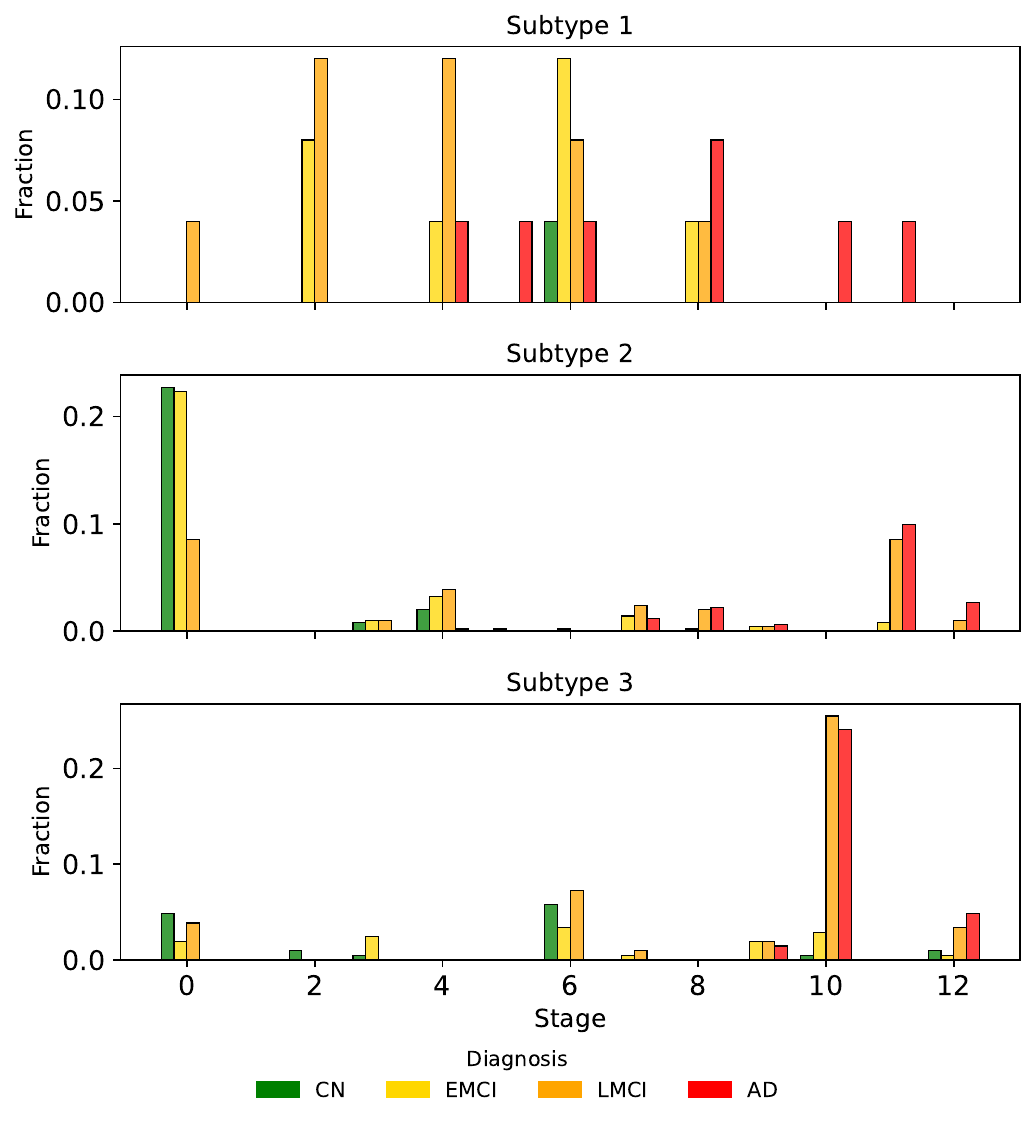}}
\end{figure}

\begin{figure}[htbp]
\floatconts
  {fig:adni_staging_sustain}
  {\vspace{-0.5cm}\caption{SuStaIn ADNI staging and subtyping
  }}
  {\includegraphics[width=0.6\linewidth]{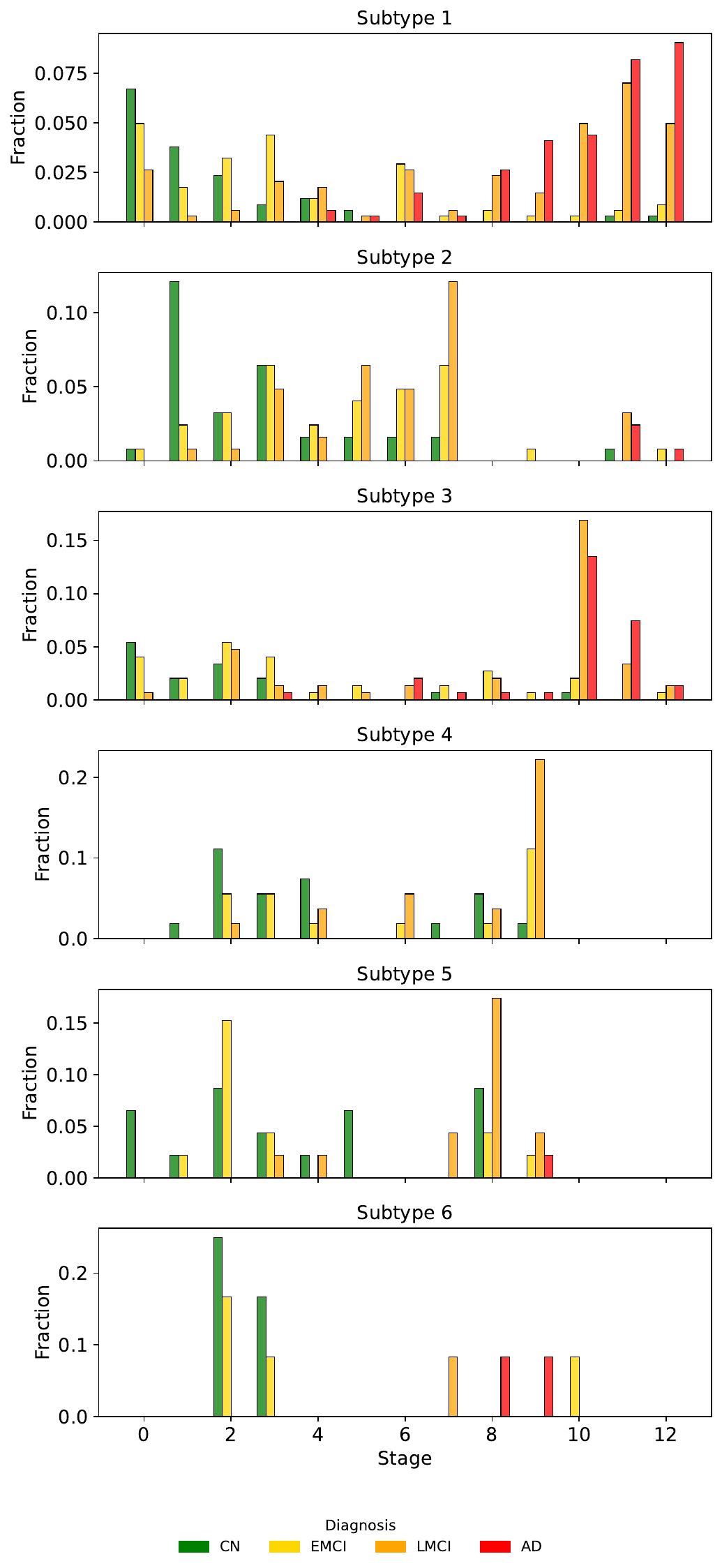}}
\end{figure}

\begin{figure*}[htbp]
 % Caption and label go in the first argument and the figure contents
 % go in the second argument
\floatconts
  {fig:adni_ordering_sustain}
  {\vspace{-0.5cm}\caption{SuStaIn ADNI ordering
  }}
  {\includegraphics[width=1.0\linewidth]{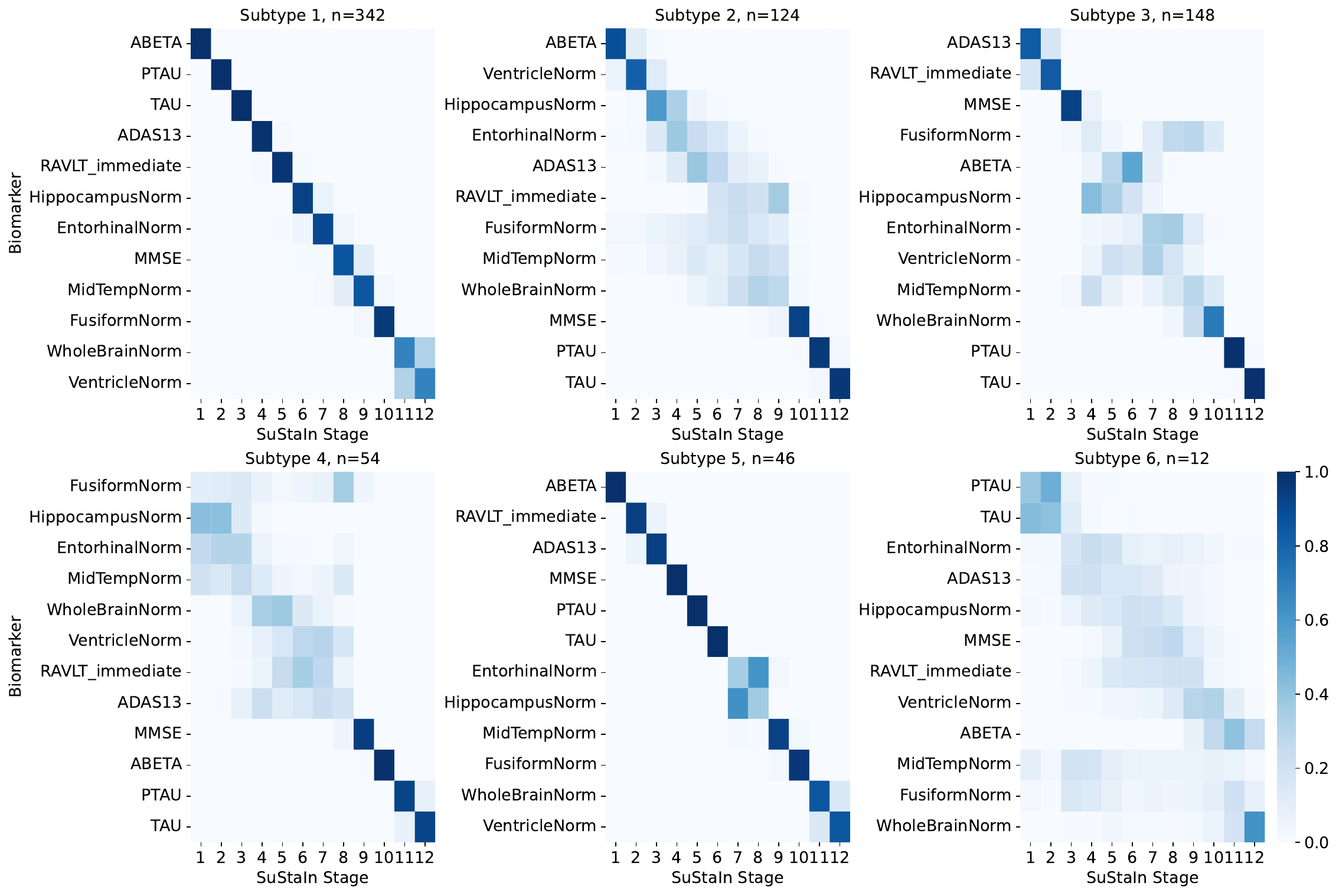}}
\end{figure*}

\clearpage 
\section{Incomplete \& High Dimensional Data}
\label{apd:high_dimensional}

\subsection{Handling of Incomplete Data}

\textsc{bebms} is designed to accommodate datasets with missing values. To achieve this, missing entries (represented as \texttt{NaN}) are systematically excluded during two key processes: (1) the initialization and iterative updating of biomarker distribution parameters, and (2) the calculation of the data likelihood. 

On the other hand, although missing data SuStaIn~\citep{estarellas2024multimodal} has been proposed, the currently available implementation of SuStaIn, whose source code was reported by \cite{estarellas2024multimodal} in the data availability section, is not able to process data with missing entries. 

\subsection{Performance on High-dimensional Data}

To assess the scalability and performance of \textsc{bebms} on high-dimensional data, we conducted a synthetic experiment. First, distribution parameters ($\boldsymbol{\theta, \phi}$) for 100 biomarkers were synthetically generated using AI. More specifically, we used the following prompt on ChatGPT5:

\begin{verbatim}
[Paste data from the JSON file into the prompt:
https://raw.githubusercontent.com/hongtaoh/bebms/refs/heads/main/4highdim_params_ucl_gmm.json]

This is for Alzheimer's disease. 

Based on this, could you give me 100 synthetic biomarkers' parameters?
\end{verbatim}

The resulting JSON can be found at \url{https://raw.githubusercontent.com/hongtaoh/bebms/refs/heads/main/high_dimensional/high_dimensional.json}. 

Based on these parameters, five distinct datasets were created following the configurations of Experiment 1 (see Appendix~\ref{apd:experiment_specifications}). The experiment was configured with $J=300$ participants and $R=0.25$, and the model was run for 3,000 MCMC iterations. All other experimental settings were identical to those described in the synthetic experiments of Section~\ref{main:model_evaluation}.

The five high-dimensional datasets are available at \url{https://github.com/hongtaoh/bebms/tree/main/high_dimensional/data}.

The results, summarized in Table~\ref{tab:high_dimensional}, indicate that \textsc{bebms} required an average processing time of approximately 10 minutes for each dataset. Considering the dimensionality of the data (100 biomarkers, 300 participants, and on average 3.4 subtypes), the model demonstrated robust performance on the ordering, subtyping, and staging tasks.

This computational speed is comparable to, though slower than, that of s-SuStaIn~\citep{tandon2024s}, which reportedly processes a dataset of 200 participants and 100 biomarkers (with 3-4 subtypes) in approximately 2 minutes. In contrast, SuStaIn GMM (without parallel start points) failed to complete processing on the 5th dataset within a 3-hour time limit. This limitation of SuStaIn aligns with the findings previously reported in the Figure 2 of~\cite{tandon2024s}.

% \subsection{High dimensional data}

% We used AI to generate distribution parameters ($\boldsymbol{\theta, \phi}$) for 100 biomarkers, and generated five datasets according to the configurations of Experiment 1 as in Appendix~\ref{apd:experiment_specifications}. We used $J=300, R=0.25$. We set the MCMC iterations to be 3,000. All other experimental configurations were the same as in synthetic experiments of Section~\ref{main:model_evaluation}.

% As Table~\ref{tab:high_dimensional} shows, \textsc{bebms} took on average around 10 minutes to process each of the five sample datasets. Considering the size of the data: 100 biomarkers and 300 participants, the performance on the ordering, subtyping and staging tasks are very good. In fact, this speed is similar to, although a little bit lower than, that of s-SuStaIn~\citep{tandon2024s}, which takes around 2 minutes to process one dataset of 200 participants with 100 biomarkers and 3-4 subtypes. 

% SuStaIn, on the other hand, was not able to process the 5th dataset in the above table within 3 hours. This result was consistent with that reported in Fig. 2 of~\cite{tandon2024s}.

\begin{table}[h!]
\centering
\caption{Comparison of runtime, ordering accuracy, subtype assignment accuracy, and healthy mean stage.}
\label{tab:high_dimensional}
\begin{tabular}{lccccc}
\hline
\textbf{Dataset} & \textbf{\# subtypes} & \textbf{Runtime (s)} & \textbf{Kendall's $\tau$} & \textbf{Subtype Acc.} & \textbf{Mean Stage (Healthy)} \\
\hline
1 & 1 & 226.58 & 0.293 & 1.000 & 0.173 \\
2 & 5 & 822.69 & 0.419 & 0.012 & 0.014 \\
3 & 4 & 719.95 & 0.378 & 0.678 & 2.164 \\
4 & 1 & 229.38 & 0.349 & 1.000 & 1.440 \\
5 & 5 & 837.29 & 0.482 & 0.016 & 1.068 \\
\hline
\textbf{Average} & 3.2 & \textbf{567.18} & \textbf{0.384} & \textbf{0.541} & \textbf{0.972} \\
\hline
\end{tabular}
\end{table}